\documentclass[journal,onecolumn]{IEEEtran}

\pdfoutput=1

\newlength{\figwidth}
\usepackage{color}
\definecolor{links}{rgb}{0.7,0,0}   %
\definecolor{urls}{rgb}{0,0,0.8}    %
\definecolor{cites}{rgb}{0,0,0.8}   %

\usepackage[colorlinks,hyperindex,linkcolor=links,citecolor=cites,urlcolor=urls]{hyperref} %

\usepackage[nosort]{cite}        
\usepackage{url} 
\usepackage[intlimits]{amsmath}
\usepackage{bbm}
\usepackage{graphicx}
\usepackage{paralist}
\usepackage{fancyref}
\usepackage[stretch=16,shrink=16,step=4]{microtype}
\usepackage{vmr-symbols-vecbold}
\usepackage{standard-macros}
\usepackage{subcaption}
\usepackage{tikz}
\usepackage{pgfplots}
\usepgfplotslibrary{fillbetween}
\usepackage{booktabs}
\usepackage[scientific-notation=true]{siunitx}

\makeatletter
\def\@IEEEinterspaceratioM{0.265}
\def\@IEEEinterspaceMINratioM{0.1651}
\def\@IEEEinterspaceMAXratioM{0.38}

\def\@IEEEinterspaceratioB{0.31}
\def\@IEEEinterspaceMINratioB{0.19}
\def\@IEEEinterspaceMAXratioB{0.38}
\@IEEEtunefonts
\makeatother
\newcommand{\trainingdata}{\boldsymbol{Z}}
\newcommand{\trainingdatasmall}{\boldsymbol{z}}
\newcommand{\supersample}{\widetilde{\boldsymbol{Z}}}
\newcommand{\supersamplesmall}{\widetilde{\boldsymbol{z}}}
\newcommand{\subsetchoice}{\boldsymbol{S}}

\newcommand{\jointdistro}{P_{W\! \supersample\! \subsetchoice }}
\newcommand{\pacBdistro}{P_{W\vert \supersample\! \subsetchoice }}

\newcommand{\trainloss}{L_{\trainingdata(\subsetchoice)}(W)}

\newcommand{\testloss}{L_{\trainingdata(\bar\subsetchoice)}(W)}
\newcommand{\trainlosswz}{L_{\trainingdatasmall(\subsetchoice)}(w)}
\newcommand{\testlosswz}{L_{\trainingdatasmall(\bar\subsetchoice)}(w)}
\newcommand{\poploss}{L_{P_Z}(W)}

\newcommand{\logQRN}{\log\frac{ \dv P_{W\! \supersample\!\subsetchoice}}{ \dv Q_{W\vert \supersample}P_{\supersample\! \subsetchoice}}}
\newcommand{\infdensop}{\imath}

\newcommand{\condinfdens}{\infdensop(W,\subsetchoice\vert \supersample)}
\newcommand{\condinfdensi}{\infdensop(W,S_i\vert \supersample)}

\begin{document}

\IEEEoverridecommandlockouts

\title{Fast-Rate Loss Bounds via Conditional Information Measures with Applications to Neural Networks}
\author{\IEEEauthorblockN{Fredrik Hellstr\"om, Giuseppe Durisi}\\
\IEEEauthorblockA{
Department of Electrical Engineering, Chalmers University of Technology, 41296 Gothenburg, Sweden\\
}}
\maketitle
\thispagestyle{plain}
\pagestyle{plain}
\begin{abstract}
We present a framework to derive bounds on the test loss of randomized learning algorithms for the case of bounded loss functions.
Drawing from Steinke \& Zakynthinou (2020), this framework leads to bounds that depend on the conditional information density between the the output hypothesis and the choice of the training set, given a larger set of data samples from which the training set is formed.
Furthermore, the bounds pertain to the average test loss as well as to its tail probability, both for the PAC-Bayesian and the single-draw settings.
If the conditional information density is bounded uniformly in the size~$n$ of the training set, our bounds decay as~$1/n$.
This is in contrast with the tail bounds involving conditional information measures available in the literature, which have a less benign $1/\sqrt{n}$ dependence. 
We demonstrate the usefulness of our tail bounds by showing that they lead to nonvacuous estimates of the test loss achievable with some neural network architectures trained on MNIST and Fashion-MNIST.
\end{abstract}

\section{Introduction}\label{sec:introduction}
In recent years, there has been a surge of interest in the use of information-theoretic techniques for bounding the loss of learning algorithms. 
While the first results of this flavor can be traced to the probably approximately correct (PAC)-Bayesian approach~\cite{mcallester98-07a,catoni07-a} (see also~\cite{guedj19-01a} for a recent review), the connection between loss bounds and classical information-theoretic measures was made explicit in the works of~\cite{russo16-05b} and~\cite{xu17-05a}, where bounds on the average population loss were derived in terms of the mutual information between the training data and the output hypothesis. 
Since then, these average loss bounds have been tightened~\cite{Bu-19-ISIT,Asadi2018,Negrea2019}.
Furthermore, the information-theoretic framework has also been successfully applied to derive tail probability bounds on the population loss~\cite{bassily18-02a,esposito19-12a}.

The information-theoretic population loss bounds in \cite{xu17-05a,Bu-19-ISIT,Asadi2018,Negrea2019,bassily18-02a,esposito19-12a} are given in terms of the training loss plus a term with a~$\sqrt{\text{IM}(n)/n}$ dependence. Here, $n$ is the number of training examples and $\text{IM}(n)$ denotes an information measure. 
This is sometimes referred to as a \textit{slow-rate} bound. 
In contrast, there exist PAC-Bayesian bounds where the dependence on $n$ is instead $\text{IM}(n)/n$ \cite{catoni07-a,mcallester13-a}, referred to as \textit{fast-rate} bounds \cite{YangRoy-2019,Grunwald-20}.\footnote{Note that our definitions of slow and fast rates coincide with the ones in \cite{Grunwald-20} only if $\text{IM}(n)$ is at most polylogarithmic in $n$. } 
Under the assumption that the information measure %
$\text{IM}(n)$ is sublinear in $n$, fast-rate bounds result in a more beneficial dependence on $n$. If this sublinearity does not hold, the bound stays constant or even increases as we increase the number of samples $n$. Thus, for all cases where the bounds are interesting, fast rates are to be preferred asymptotically.

The purpose of this paper is to derive and evaluate fast-rate information-theoretic tail bounds on the test loss for the \textit{random-subset setting} introduced in~\cite{steinke20-a}. %
In this setting,~$2n$ training samples~$\supersample=(\tilde Z_1,\dots,\tilde Z_{2n})$ are available, with all entries of~$\supersample$ being drawn independently from some distribution~$P_Z$ on an instance space~$\mathcal{Z}$. However, only a randomly selected subset of cardinality~$n$ is actually used for training. It is selected as follows: 
let~$\subsetchoice=(S_1,\dots,S_n)$ be an~$n$-dimensional random vector, the elements of which are drawn independently from a~$\mathrm{Bern}(1/2)$ distribution and are independent of~$\supersample$. 
Then, for~$i=1,\dots,n$, the~$i$th training sample in~$\trainingdata(\subsetchoice)$ is~$Z_i(S_i)=\tilde Z_{i+S_in}$. 
Based on this training set, a hypothesis~$W\in \mathcal{W}$ is chosen through a randomized learning algorithm $P_{W\vert\supersample\!\subsetchoice}=P_{W\vert\trainingdata(\subsetchoice)}$, which is a conditional distribution on $\mathcal{W}$ given $(\supersample,\subsetchoice)$ that gives rise to the Markov property $(\supersample,\subsetchoice)-\trainingdata(\subsetchoice)-W$. 
Let~$L_{\trainingdata(\subsetchoice)}(W)=\tfrac{1}{n}\sum_{i=1}^n\ell(W,Z_i(S_i))$ denote the training loss, where $\ell(\cdot,\cdot)$ is a loss function, which throughout this paper is assumed to be restricted to $[0,1]$. 
Furthermore, let~$\bar \subsetchoice$ denote the modulo-2 complement of~$\subsetchoice$.
Then~$L_{\trainingdata(\bar\subsetchoice)}(W)$ can be interpreted as a test loss, since~$W$ is conditionally independent of~$\trainingdata(\bar \subsetchoice)$ given $\trainingdata( \subsetchoice)$. 
Note that the average over~$(\supersample,\subsetchoice)$ of the test loss is the population loss~$L_{P_Z}(W)=\Exop_{P_{\supersample\!\subsetchoice}}[\testloss]=\Exop_{P_Z}[\ell(W,Z)]$.
For this setting, bounds on the average population loss are derived in~\cite{steinke20-a} in terms of the conditional mutual information (CMI) $I(W;\subsetchoice\vert\supersample)$ between the chosen hypothesis~$W$ and the random vector~$\subsetchoice$ given the set $\supersample$. Bounds for the random-subset setting are always finite, since~$I(W;\subsetchoice\vert\supersample)$ is never larger than~$n$ bits.  
In contrast, the bounds obtained in~\cite{xu17-05a} depend on the mutual information~$I(W;\trainingdata)$, a quantity that can be unbounded if~$W$ reveals too much about the training set~$\trainingdata$. 

The following bounds, the second of which is a fast-rate bound, are derived in~\cite[Thm.~2]{steinke20-a}:
\begin{align}\label{eq:steinke_slow}
  &\Ex{\jointdistro}{\poploss} \leq \Ex{\jointdistro}{\trainloss}   +    \sqrt{\frac{2I(W;\subsetchoice\vert \supersample) }{n}} \\ \label{eq:steinke_fast}
    & \Ex{\jointdistro}{\poploss} \leq 2 \Ex{\jointdistro}{\trainloss} + \frac{3I(W;\subsetchoice\vert \supersample) }{n}.
\end{align}
The price for the fast rate in~\eqref{eq:steinke_fast} is that the training loss that is added to the~$n$-dependent term is multiplied by a constant larger than~$1$. 
We note that the results in~\cite[Thm.~2]{steinke20-a} pertain only to the average population loss: no tail bounds are provided. 

The slow-rate average bound~\eqref{eq:steinke_slow} can be extended to the PAC-Bayesian and single-draw settings. In Appendix A, we derive the following two bounds.\footnote{Similar bounds were recently reported in~\cite[Cor. 6-7]{hellstrom-20b}. However, one step in the proofs, involving the optimization over a parameter $\lambda$, is incorrect.} %
With probability at least~$1-\delta$ under~$P_{\supersample\!\subsetchoice}$,
\begin{equation}\label{eq:hellstrom_slow_pacb}
 \Exop_{P_{W\vert \supersample\! \subsetchoice }} \lefto[\testloss\right]   \leq \Exop_{P_{W\vert \supersample\! \subsetchoice }} \lefto[\trainloss\right]
 + \sqrt{\frac{2}{n-1}\left(\relent{P_{W\vert \supersample\! \subsetchoice }}{P_{W\vert \supersample } } +\log \frac{\sqrt{n}}{\delta}\right)}.
\end{equation}
Furthermore, with probability at least~$1-\delta$ under~$\jointdistro$,
\begin{equation}\label{eq:hellstrom_slow_sd}
\testloss   \!\leq\! \trainloss \!+\!  \sqrt{{\frac{2}{n-1}\left( \condinfdens\!+\!\log \frac{\sqrt{n}}{\delta}\right)} } .
\end{equation}%
Here, the conditional information density~$\condinfdens$ between~$W$ and~$\subsetchoice$ given~$\supersample$ is defined as~$\condinfdens=\log \tfrac{\dv \jointdistro}{\dv P_{W\vert \supersample}P_{\supersample\!\subsetchoice}}$. %
In practical applications, $P_{W\vert \supersample}$ is difficult to compute, since marginalizing~$P_{\subsetchoice}P_{W\vert \supersample\!\subsetchoice}$ over~$\subsetchoice$ involves running the learning algorithm~$2^n$ times. Typically, this means that~$\condinfdens$ cannot be evaluated.
Therefore, in this paper we will replace~$\condinfdens$ with the proxy~$\log \tfrac{\dv \jointdistro}{\dv Q_{W\vert \supersample}P_{\supersample\!\subsetchoice}}$. Here,~$Q_{W\vert \supersample}$ is a suitably chosen auxiliary distribution (\emph{prior}) used in place of the intractable, true marginal. 
While the bounds in~\eqref{eq:hellstrom_slow_pacb} and~\eqref{eq:hellstrom_slow_sd} pertain to the test loss instead of the population loss, one can obtain population loss bounds by adding a penalty term to~\eqref{eq:hellstrom_slow_pacb} and~\eqref{eq:hellstrom_slow_sd}, as shown in~\cite[Thm.~2]{hellstrom-20b}. 
However, when comparing bounds to the empirical performance of an algorithm, the population loss is unknown. 
Thus, in practice, one has to resort to evaluating a test loss.
\paragraph*{Contributions}
In this paper, we derive fast-rate versions of \eqref{eq:hellstrom_slow_pacb} and \eqref{eq:hellstrom_slow_sd}, thereby extending the fast-rate average loss bound in~\cite{steinke20-a} to the PAC-Bayesian and single-draw settings. We then use the resulting PAC-Bayesian and single-draw bounds to characterize the test loss of neural networks (NN) used to classify images from the MNIST and Fashion-MNIST data sets. To obtain nonvacuous bounds for NNs, it is crucial to choose a data-dependent prior~\cite{dziugaite-20}. The random-subset setting provides a natural way to do this by choosing $Q_{W\vert\supersample}$ as an approximation of $P_{W\vert\supersample}$.
The single-draw bounds that we present can be applied to deterministic NNs trained through stochastic gradient descent (SGD) with Gaussian noise added to the final weights, whereas the PAC-Bayesian bounds apply only to randomized NNs, whose weights are drawn from a Gaussian distribution each time the network is used. 
For the same setup, we also evaluate the slow-rate PAC-Bayesian and single-draw bounds from~\eqref{eq:hellstrom_slow_pacb} and~\eqref{eq:hellstrom_slow_sd}. Our numerical results reveal that both the slow-rate and fast-rate bounds are nonvacuous, and in line with previously reported results for similar setups~\cite{dziugaite-20}. While our bounds improve with $n$, illustrating that the information measures are sublinear in $n$, the difference between the slow-rate and fast-rate bounds is minor. This indicates that, for our choice of learning algorithm $P_{W\vert\supersample\!\subsetchoice}$ and prior $Q_{W\vert\supersample}$, the sublinearity is mild.
\section{Fast-Rate Random-Subset Bounds}\label{sec:fastratebounds}
We start by presenting an exponential inequality from which several test-loss bounds can be derived. 
This result and its proof illustrate how to combine the exponential-inequality approach from~\cite{hellstrom-20b} with fast-rate derivations, like those presented in~\cite[Thm.~2]{mcallester13-a} and~\cite[Thm.~2.(3)]{steinke20-a}. In order to avoid measurability issues, we will assume throughout this paper that the supports of $Q_{W\vert \supersample}P_{\supersample}P_{\subsetchoice}$ and $\jointdistro=P_{W\vert \supersample\!\subsetchoice}P_{\supersample}P_{\subsetchoice}$ coincide.
\begin{thm}\label{thm:exp_inequality}
Consider the random-subset setting introduced in Section~\ref{sec:introduction}. Let $W\in \mathcal{W}$ be distributed according to $P_{W\vert \trainingdata(\subsetchoice)}$. 
Let~$\lambda,\gamma>0$ be constants such that $\lambda(1-\gamma)+(e^{\lambda}-1-\lambda)(1+\gamma^2)\leq 0$. Furthermore, let $Q_{W\vert\supersample}$ be an arbitrary conditional prior.
Then, the following holds:
\begin{equation}\label{eq:exp_ineq}
 \Exop_{P_{W\! \supersample\! \subsetchoice }} \biggo[\exp\biggo(\lambda n\left( L_{\trainingdata(\bar\subsetchoice)}(W)-\gamma  L_{\trainingdata(\subsetchoice)}(W)\right)-\logQRN \bigg)\bigg]   \leq 1.
\end{equation}
\end{thm}
\begin{IEEEproof}
We begin by proving an exponential inequality for a binary random variable~$X$ satisfying~$P(X=a)=P(X=b)=1/2$ where~$a,b\in [0,1]$. Let~$\bar X =b$ if $X=a$ and $\bar{X}=a$ if $X=b$. 
Finally, let $c=e^{\lambda}-1-\lambda$. Then,
\begin{equation}
\Ex{}{e^{\lambda\left(X-\gamma\bar X\right)}} \leq \Ex{}{1+\lambda\left(X-\gamma\bar X\right)+c\left(X-\gamma\bar X\right)^2}
=1+\frac{\lambda(1-\gamma)}{2}\left(a+b\right)+\frac{c}{2}\left(a-\gamma b\right)^2+\frac{c}{2}\left(b-\gamma a\right)^2.
\end{equation}
Here, the first inequality follows because $e^y \leq 1 +y+ cy^2/{\lambda^2}$  for all $y\leq \lambda$.
Expanding the squares and removing negative terms, we find that
\begin{align}
\Ex{}{e^{\lambda\left(X-\gamma\bar X\right)}} %
&\leq 1+\lambda(1-\gamma)+(e^{\lambda}-1-\lambda)(1+\gamma^2)\leq 1,\label{eq:exp-ineq-generic}
\end{align}
where the second inequality follows from our assumption on $\lambda,\gamma$.
Let $Q_{W\!\supersample}=Q_{W\vert \supersample} P_{\supersample}$, and apply~\eqref{eq:exp-ineq-generic} with $X=\ell(w,Z_i(\bar S_i))$ and $\bar X=\ell(w,Z_i(S_i))$ for some fixed $w$ and~$\supersamplesmall$.
It follows that 
\begin{equation}
\Ex{Q_{W\!\supersample}P_{\subsetchoice}}{e^{\lambda n\bigl(\testloss-\gamma\trainloss\bigr)}} = \Ex{Q_{W\!\supersample}}{\prod_{i=1}^n\Ex{P_{S_i}}{{e^{\lambda\bigl(\ell(W,Z_i(\bar S_i))-\gamma\ell(W,Z_i(S_i))\bigr)}}} }\leq 1.
\end{equation}
The desired result now follows after a change of measure to~$\jointdistro$~\cite[Prop.~17.1]{polyanskiy19-a}.
\end{IEEEproof}

Note that the exponential function in \eqref{eq:exp_ineq} depends linearly on the test loss. In contrast, the exponential inequality in~\eqref{eq:slow_rate_exp_ineq} in Appendix~\ref{sec:app_A}, used to establish \eqref{eq:hellstrom_slow_pacb} and \eqref{eq:hellstrom_slow_sd}, depends quadratically on the test loss. This difference explains why Theorem~\ref{thm:exp_inequality} allows for the derivation of fast-rate bounds, whereas~\eqref{eq:slow_rate_exp_ineq} unavoidably leads to slow-rate bounds.

By simple applications of Jensen's and Markov's inequalities, the exponential inequality~\eqref{eq:exp_ineq} can be used to derive bounds on the population loss and on the test loss. We present these bounds in the following corollary.
\begin{cor}\label{cor:bounds}
Consider the setting of Theorem~\ref{thm:exp_inequality}. Then, the average population loss is bounded by
\begin{equation}\label{eq:avg_bound}
    \Ex{\jointdistro}{\poploss} \leq \gamma \Ex{\jointdistro}{\trainloss} + \frac{\Ex{P_{\supersample\!\subsetchoice}}{\relent{P_{W\vert \supersample\! \subsetchoice }}{Q_{W\vert \supersample } }}}{\lambda n}.
\end{equation}
Furthermore, with probability at least~$1-\delta$ over~$P_{\supersample\!\subsetchoice}$, the PAC-Bayesian test loss is bounded by
\begin{equation}\label{eq:pacb_bound}
    \Ex{\pacBdistro}{\testloss} \leq  \gamma\Ex{\pacBdistro}{\trainloss} + \frac{\left(\relent{\pacBdistro}{Q_{W\vert \supersample}}+\log \frac{1}{\delta}\right)}{\lambda n}.
\end{equation}
Finally, with probability at least~$1-\delta$ over~$\jointdistro$, the single-draw test loss is bounded by
\begin{equation}\label{eq:sd_bound}
    {\testloss} \leq  \gamma{\trainloss} + \frac{\left(\logQRN+\log \frac{1}{\delta}\right)}{\lambda n}.
\end{equation}
\end{cor}
\begin{IEEEproof}
We begin by applying Jensen's inequality to~\eqref{eq:exp_ineq} to move the expectation inside the exponential. 
We then obtain~\eqref{eq:avg_bound} by taking the logarithm of both sides and reorganizing terms.

To derive~\eqref{eq:pacb_bound}, we first apply Jensen's inequality in~\eqref{eq:exp_ineq}, this time only with respect only~$P_{W\vert\supersample\!\subsetchoice}$, to get
\begin{equation}\label{eq:pacb_derive_before_markov}
 \Exop_{P_{\supersample\! \subsetchoice }} \biggo[\exp\biggo(\Ex{P_{W\vert\supersample\!\subsetchoice}}{\lambda n\left( L_{\trainingdata(\bar\subsetchoice)}(W)- \gamma L_{\trainingdata(\subsetchoice)}(W)\right)}-\relent{\pacBdistro}{Q_{W\vert \supersample}} \bigg)\bigg] \leq 1.
\end{equation}
We now use Markov's inequality in the following form. Let~$U\distas P_U$ be a nonnegative random variable satisfying~$\Ex{}{U}\leq 1$. 
Then,
\begin{equation}\label{eq:markov_inequality_Pu}
P_U[U\leq 1/\delta]\geq 1- \Ex{ }{U}\delta\geq 1- \delta.
\end{equation}
Applying~\eqref{eq:markov_inequality_Pu} to~\eqref{eq:pacb_derive_before_markov} we find that, with probability at least~$1-\delta$ under~$P_{\supersample\!\subsetchoice}$,
\begin{equation}
\exp\biggo(\Ex{P_{W\vert\supersample\!\subsetchoice}}{\lambda n\left(L_{\trainingdata(\bar\subsetchoice)}(W)-\gamma  L_{\trainingdata(\subsetchoice)}(W)\right)}-\relent{\pacBdistro}{Q_{W\vert \supersample}} \bigg)\leq \frac{1}{\delta}.
\end{equation}
Taking the logarithm and reorganizing terms, we obtain~\eqref{eq:pacb_bound}.

Finally, to derive~\eqref{eq:sd_bound}, we directly apply~\eqref{eq:markov_inequality_Pu} to~\eqref{eq:exp_ineq} to conclude that, with probability at least~$1-\delta$ under~$\jointdistro$,
\begin{equation}
\exp\lefto(\lambda n\lefto(L_{\trainingdata(\bar\subsetchoice)}(W)-\gamma  L_{\trainingdata(\subsetchoice)}(W)\right)-\logQRN \right) \leq \frac{1}{\delta}.
\end{equation}
The desired bound~\eqref{eq:sd_bound} follows after taking the logarithm and reorganizing terms.
\end{IEEEproof}
The bounds in~\eqref{eq:pacb_bound} and~\eqref{eq:sd_bound} are data-dependent, i.e., they depend on the specific instances of~$\supersample$ and~$\subsetchoice$. 
They can be turned into data-independent bounds that are functions of the average of the information measures appearing in~\eqref{eq:pacb_bound} and~\eqref{eq:sd_bound}, at the cost of a less benign polynomial dependence on the confidence parameter~$\delta$.
Alternatively, one can obtain bounds that have a more benign dependence on $\delta$ if one allows the bounds to depend on sufficiently high moments of the information measures appearing in~\eqref{eq:pacb_bound} and~\eqref{eq:sd_bound}, or if one replaces these measures by quantities such as conditional maximal leakage or conditional~$\alpha$-divergence.
See~\cite{hellstrom-20b} for further discussion.

Setting~$Q_{W\vert \supersample}=P_{W\vert\supersample}$, $\gamma=2$ and $\lambda=1/3$ in~\eqref{eq:avg_bound}, we recover the CMI bound in~\cite{steinke20-a}. %
As illustrated in Corollary~\ref{cor:samplewise_bound} below, for the special case $Q_{W\vert\supersample}=P_{W\vert\supersample}$, the bound on the average population loss in~\eqref{eq:avg_bound} can be tightened by replacing the CMI $\Ex{P_{\supersample\!\subsetchoice} }{\relent{\pacBdistro}{P_{W\vert\supersample}}}=I(W;\subsetchoice\vert\supersample)$ with a sum of samplewise CMIs $I(W;S_i\vert\supersample)$. 
\begin{cor}\label{cor:samplewise_bound}
Consider the setting of Theorem~\ref{thm:exp_inequality}, with the additional assumption that $Q_{W\vert\supersample}=P_{W\vert \supersample}$. Then, the average population loss is bounded by
\begin{equation}\label{eq:samplewise_avg_bound}
    \Ex{\jointdistro}{\poploss} \leq \gamma \Ex{\jointdistro}{\trainloss} + \sum_{i=1}^n\frac{I(W;S_i\vert \supersample)}{\lambda n}.
\end{equation}
\end{cor}
\begin{IEEEproof}
Consider a fixed~$w\in\mathcal{W}$ and~$\supersamplesmall\in\mathcal{Z}^{2n}$. By~\eqref{eq:exp-ineq-generic},
\begin{equation}
\Ex{P_{S_i}}{e^{\lambda \left(\ell(w,Z_i(\bar S_i)) - \gamma\ell(w,Z_i(S_i)) \right)} } \leq 1.
\end{equation}
Let~$P_{S_i\vert w\supersamplesmall}$ denote $P_{S_i\vert W=w,\supersample=\supersamplesmall}$ for some fixed $w,\supersamplesmall$. By changing measure to $P_{S_i\vert w\supersamplesmall}$ we obtain
\begin{equation}
\Ex{P_{S_i}}{ e^{\lambda \left(\ell(w,Z_i(\bar S_i)) - \gamma\ell(w,Z_i(S_i)) \right)} } = \Ex{P_{S_i\vert w\supersamplesmall}}{e^{\lambda \left(\ell(w,Z_i(\bar S_i)) - \gamma\ell(w,Z_i(S_i)) \right) - \log\frac{\dv P_{S_i\vert w\supersamplesmall}}{\dv P_{S_i}}} } \leq 1.
\end{equation}
Moving the expectation inside the exponential through the use of Jensen's inequality and taking the logarithm, we obtain
\begin{equation}
 \Ex{P_{S_i\vert w\supersamplesmall}}{\ell(w,Z_i(\bar S_i))} 
 \leq \gamma\Ex{P_{S_i\vert w\supersamplesmall}}{\ell(w,Z_i(S_i))}  + \frac{1}{\lambda}\Ex{P_{S_i\vert w\supersamplesmall}}{\log\frac{\dv P_{S_i\vert w\supersamplesmall}}{\dv P_{S_i}}}
 = \gamma\Ex{P_{S_i\vert w\supersamplesmall}}{\ell(w,Z_i(S_i))}  + \frac{\relent{P_{S_i\vert w\supersamplesmall}}{P_{S_i}}}{\lambda}.\label{eq:deriv_samplewise_i}
\end{equation}
The desired result follows by noting that
\begin{IEEEeqnarray}{C}\label{eq:deriv_samplewise_decomp}
\Ex{\jointdistro}{\poploss}=\Ex{P_{W\!\supersample}}{ \sum_{i=1}^n \Ex{P_{S_i\vert W\supersample}}{\frac{\ell(W,Z_i(\bar S_i))}{n} } }
\end{IEEEeqnarray}
and applying~\eqref{eq:deriv_samplewise_i} to each term in the sum in~\eqref{eq:deriv_samplewise_decomp}.
\end{IEEEproof}

For the so-called interpolating setting, where~$\trainloss=0$, one can obtain a different exponential inequality than the one reported in Theorem~\ref{thm:exp_inequality}, under the additional assumption that $Q_{W\vert \supersample}=P_{W\vert \supersample}$. This leads to tighter bounds than the ones in Corollary~\ref{cor:bounds}.
Specifically, in these alternative bounds, the factor~$\lambda$ can be set to~$\log 2\approx 0.69$. In contrast, any $\lambda$ in Theorem~\ref{thm:exp_inequality}, regardless of the value of $\gamma$, must satisfy $\lambda^2-4(e^\lambda-1)(e^\lambda-1-\lambda)\geq 0$, which implies $\lambda< 0.37$. %

We begin by proving the following exponential inequality, the derivation of which is similar to part of the proof of the fast-rate bound for the interpolating setting given in~\cite{steinke20-a}. %

\begin{thm}\label{thm:interp_exp_ineq}
Consider the random-subset setting introduced in Section~\ref{sec:introduction}. Let $W\in \mathcal{W}$ be distributed according to $P_{W\vert \trainingdata(\subsetchoice)}$, and  assume that~$\trainloss=0$ for $W\distas P_{W\vert \trainingdata(\subsetchoice)}$. %
Then,
\begin{equation}\label{eq:thm_realizable_exp_ineq}
\Ex{\jointdistro}{\exp\lefto(n\log2\cdot \testloss - \condinfdens\right)} \leq 1.
\end{equation}
\end{thm}
\begin{IEEEproof}
Let~$\lambda,\gamma>0$. Then, %
\begin{equation}
\Exop_{P_{W\!\supersample\!\subsetchoice}}\biggo[\prod_{i=1}^n\biggo( \frac{1}{2} e^{\lambda \ell(W, Z_i(\bar S_i)) - \gamma \ell(W, Z_i( S_i)) }
+\frac{1}{2} e^{\lambda \ell(W, Z_i( S_i)) - \gamma \ell(W, Z_i(\bar S_i)) } \bigg)\bigg] 
= \Exop_{P_{W\!\supersample}P_{\subsetchoice}}\biggo[ \prod_{i=1}^ne^{\lambda \ell(W, Z_i(\bar S_i)) - \gamma \ell(W, Z_i(S_i))  } \bigg].\label{eq:interp_deriv_primes}
\end{equation}
It follows from~\eqref{eq:interp_deriv_primes} that
\begin{equation}
\Ex{P_{W\!\supersample}P_{\subsetchoice}}{e^{n\left(\lambda \testloss-\gamma\trainloss \right)}}    
= \Exop_{P_{W\!\supersample\!\subsetchoice}}\biggo[\prod_{i=1}^n\bigg( \frac{1}{2} e^{\lambda \ell(W, Z_i(\bar S_i)) - \gamma \ell(W, Z_i( S_i)) }
+\frac{1}{2} e^{\lambda \ell(W, Z_i( S_i)) - \gamma \ell(W, Z_i(\bar S_i)) } \bigg)\bigg].
\end{equation}
We now change measure to~$\jointdistro$ to conclude that
\begin{equation}\label{eq:interp_deriv_before_paramset}
\Ex{\jointdistro}{e^{n\left(\lambda \testloss-\gamma\trainloss \right) - \condinfdens}}    
=\Exop_{P_{W\!\supersample\!\subsetchoice}}\biggo[\prod_{i=1}^n\bigg( \frac{1}{2} e^{\lambda \ell(W, Z_i(\bar S_i)) - \gamma \ell(W, Z_i( S_i)) }
+\frac{1}{2} e^{\lambda \ell(W, Z_i( S_i)) - \gamma \ell(W, Z_i(\bar S_i)) }\bigg) \bigg].
\end{equation}
We now use the interpolating assumption and set~$\lambda=\log 2$. If $\ell(W, Z_i(\bar S_i))=0$, \eqref{eq:thm_realizable_exp_ineq} holds trivially for every $\gamma$. If $\ell(W, Z_i(\bar S_i))> 0$, we let~$\gamma\rightarrow\infty$. This, together with the assumption that~$\ell(W, Z_i(\bar S_i)) \in [0,1]$, implies \eqref{eq:thm_realizable_exp_ineq}. 
\end{IEEEproof}

Using Theorem~\ref{thm:interp_exp_ineq}, we can derive bounds that are analogous to those in Corollary~\ref{cor:bounds}. 
We present these bounds below without proof, since they can be established following steps similar to the ones used to prove Corollary~\ref{cor:bounds}.

\begin{cor}\label{cor:interpolating_bounds}
Consider the setting of Theorem~\ref{thm:interp_exp_ineq}. 
Then, the average population loss is bounded by\footnote{Since $I(W;\subsetchoice\vert \supersample)\leq \log 2^n$ for all distributions, the constant $\log 2$ cannot be improved.}
\begin{equation}\label{eq:interp_average_bound}
    \Ex{\jointdistro}{\poploss} \leq  \frac{I(W;\subsetchoice\vert \supersample)}{n \log 2}.
\end{equation}
Furthermore, with probability at least~$1-\delta$ over~$P_{\supersample\!\subsetchoice}$, the PAC-Bayesian population loss is bounded by
\begin{equation}
    \Ex{\pacBdistro}{\testloss} \leq   \frac{ \relent{\pacBdistro}{P_{W\vert \supersample}}+\log \frac{1}{\delta}}{n\log 2}.
\end{equation}
Finally, with probability at least~$1-\delta$ over~$\jointdistro$, the single-draw population loss is bounded by
\begin{equation}
    {\testloss} \leq  \frac{ \condinfdens+\log \frac{1}{\delta}}{n \log 2}.
\end{equation}
\end{cor}
To conclude, we present a samplewise average bound for the interpolating setting, which can be shown to be tighter than \eqref{eq:interp_average_bound} by using the chain rule for mutual information and the independence of the $S_i$, as in~\cite[Rem. 3.5]{Haghifam2020}.
\begin{cor}
Consider the setting of Theorem~\ref{thm:interp_exp_ineq}. Then, the average population loss is bounded by
\begin{equation}
    \Ex{\jointdistro}{\poploss} \leq  \sum_{i=1}^n\frac{I(W;S_i\vert \supersample)}{n\log 2}.
\end{equation}
\end{cor}

\begin{IEEEproof}
Let~$\lambda,\gamma>0$. For all~$i$, by arguing as in \eqref{eq:interp_deriv_primes}---\eqref{eq:interp_deriv_before_paramset},
\begin{equation}\label{eq:interp_deriv_before_paramset_samplewise}
\Ex{P_{W\!\supersample\! S_i}}{ e^{\lambda \ell(W, Z_i(\bar S_i)) - \gamma \ell(W, Z_i(S_i))-\condinfdensi } }    
= \Exop_{P_{W\!\supersample\!\subsetchoice}}\biggo[\bigg( \frac{1}{2} e^{\lambda \ell(W, Z_i(\bar S_i)) - \gamma \ell(W, Z_i( S_i)) }+\frac{1}{2} e^{\lambda \ell(W, Z_i( S_i)) - \gamma \ell(W, Z_i(\bar S_i)) } \bigg)\bigg]. 
\end{equation}
Here,~$\condinfdensi=\log \frac{\dv P_{W\!\supersample\!S_i}}{\dv P_{W\!\supersample}P_{S_i}}$. %
We now use the interpolating assumption and set~$\lambda=\log 2$. 
If $\ell(W, Z_i(\bar S_i))> 0$, we let~$\gamma\rightarrow\infty$. This, together with the assumption that~$\ell(W, Z_i(\bar S_i)) \in [0,1]$, implies that the right-hand side of~\eqref{eq:interp_deriv_before_paramset_samplewise} is no larger than 1. If $\ell(W, Z_i(\bar S_i))=0$, this holds trivially for every $\gamma$.
Thus,
\begin{equation}\label{eq:interp_deriv_exp_ineq_samplewise}
\Ex{P_{W\!\supersample\! S_i}}{ e^{\log 2\cdot \ell(W, Z_i(\bar S_i))-\condinfdensi  } }    
\leq 1.
\end{equation}
By using Jensen's inequality to move the expectation inside the exponential and reorganizing the resulting inequality, we obtain
\begin{equation}\label{eq:interp_deriv_samplewise_for_one_i}
\Ex{P_{W\!\supersample\! S_i}}{ \ell(W, Z_i(\bar S_i)) } 
\leq \frac{I(W;S_i\vert \supersample)}{\log 2}.
\end{equation}
The result now follows because
\begin{equation}
\Ex{P_{W\!\supersample\! \subsetchoice}}{\poploss }=\Ex{P_{W\!\supersample\! \subsetchoice}}{\frac{1}{n}\sum_{i=1}^n \ell(W, Z_i(\bar S_i)) }.%
\end{equation}
\end{IEEEproof}
\section{Experiments}\label{sec:experiments}
In this section, we numerically evaluate the bounds in~\eqref{eq:hellstrom_slow_pacb},~\eqref{eq:hellstrom_slow_sd},~\eqref{eq:pacb_bound} and~\eqref{eq:sd_bound} for some NNs. Specifically, we consider the convolutional network LeNet-5 and a fully connected NN with two hidden layers of width 600, trained on either MNIST or Fashion-MNIST using SGD. A full description of the network architectures is given in Appendix \ref{app:exp_architectures}. We set the loss function to be the classification error. For the setups considered, the PAC-Bayesian and single-draw bounds are numerically indistinguishable, so we only present the PAC-Bayesian ones. To evaluate the bounds, we want to choose $\lambda$ as large as possible and $\gamma$ as small as possible. We will use~$\lambda = 1/2.98$ and~$\gamma = 1.795$.

We set the posterior~$P_{W\vert \supersample\!\subsetchoice}$ to be $\normal(W\mid \boldsymbol{\mu}_1,\sigma^2_1\matI_d)$, where $\boldsymbol{\mu}_1$ contains the $d$ NN weights found by SGD on the training set $\trainingdata(\subsetchoice)$, a randomly chosen subset of $n$ samples from the $2n$ available in $\supersample$. 
The parameter $\sigma_1^2$ is chosen as large as possible, to some finite precision, so that the training accuracy of the stochastic NN with weights drawn from $\normal(W\mid \boldsymbol{\mu}_1,\sigma^2_1\matI_d)$ differs by at most some threshold from the training accuracy of the deterministic NN with weights $\boldsymbol{\mu}_1$. The thresholds used for the different setups are specified in Appendix~\ref{app:exp_training}. %

The prior $P_{W\vert \supersample}$ can in principle be computed by averaging over all $2^n$ possible values of $\subsetchoice$. While such an exact computation is prohibitively expensive, this indicates a principled way to choose a prior by approximately performing this procedure. To choose the prior, we therefore proceed as follows. First, we form $10$ subsets of $\supersample$, and train an NN with SGD on each, denoting the average of the output weights as $\boldsymbol{\mu}_2$. We then find $\tilde\sigma_2$ so that the accuracy on $\supersample$ for the stochastic NN with weights drawn from $\normal(W\mid\boldsymbol{\mu}_2,\tilde\sigma_2^2\matI_d)$ and the deterministic NN with weights $\boldsymbol{\mu}_2$ differs by at most the specified threshold. On the basis of $\tilde\sigma_2$, we create a set of candidate values for $\sigma_2$. We then set $Q_{W\vert \supersample}=\normal(W\mid\boldsymbol{\mu}_2,\sigma_2^2\matI_d)$, where $\sigma_2$ is chosen to minimize the bound, typically leading to $\sigma_1=\sigma_2$. For this final bound to be valid, we take a union bound over the set of candidate values. More details on the training procedure and choice of $\sigma_1$, $\sigma_2$ are given in Appendix \ref{app:exp_training}.

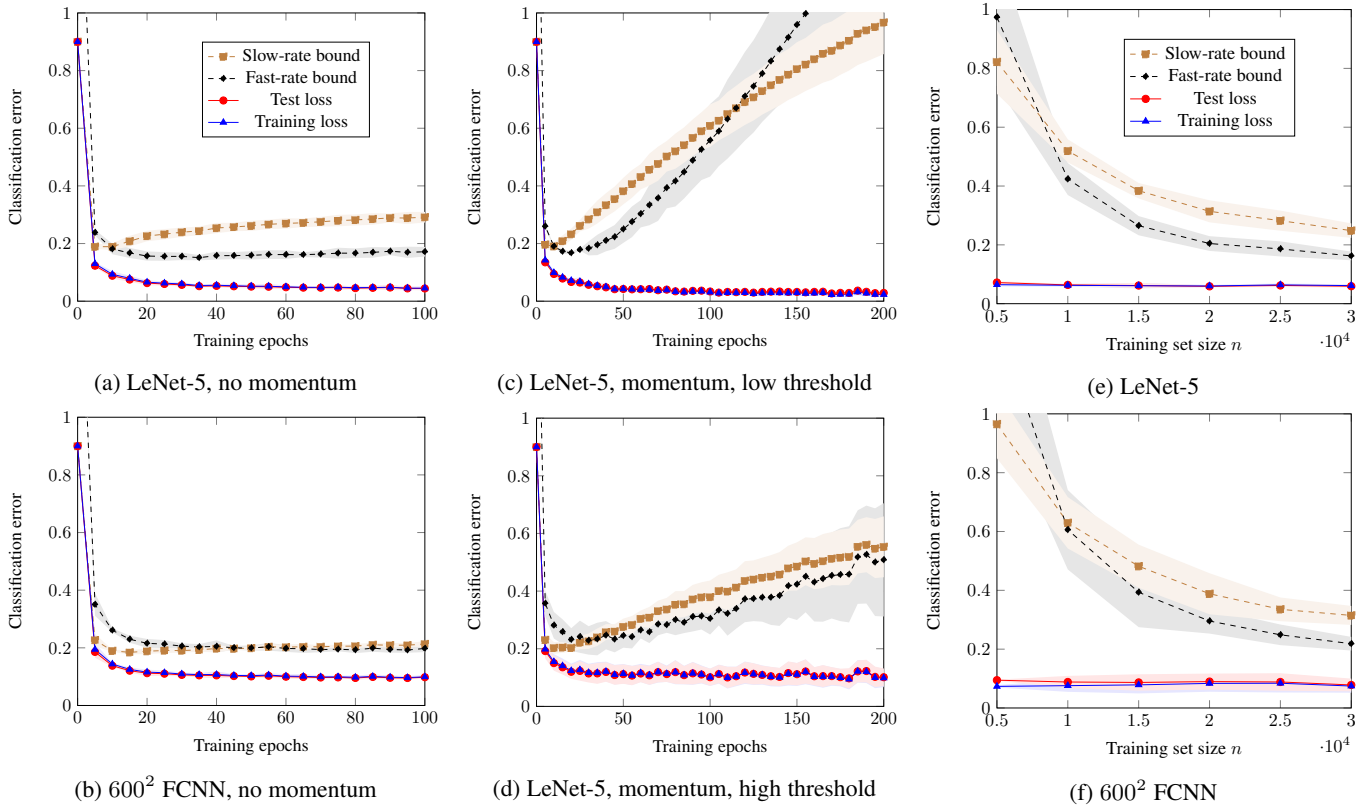
\begin{figure}[t]
\renewcommand{\thesubfigure}{a}
\begin{subfigure}{.33\textwidth}
    \centering
    \resizebox{.99\textwidth}{!}{%
    \begin{tikzpicture}
    \begin{axis}[ymin=0, ymax=1.0, legend style={at={(0.60,0.90)},anchor=north},xmin=0,xmax=100.0,
         xlabel=Training epochs,
         ylabel=Classification error]
    \addplot[brown,dashed,mark=square*] table [y=slowavg, x=E]{LeNet_MN.dat};
    \addlegendentry{Slow-rate bound}
    \addplot[black,dashed,mark=diamond*] table [y=fastavg, x=E]{LeNet_MN.dat};
    \addlegendentry{Fast-rate bound}
    \addplot[red,solid,mark=otimes*] table [y=testavg, x=E]{LeNet_MN.dat};
    \addlegendentry{Test loss}
    \addplot[blue,solid,mark=triangle*] table [y=trainavg, x=E]{LeNet_MN.dat};
    \addlegendentry{Training loss}
    
    \addplot [name path=upper,draw=none] table[x=E,y expr=\thisrow{trainavg}+2*\thisrow{trainstd}] {LeNet_MN.dat};
\addplot [name path=lower,draw=none] table[x=E,y expr=\thisrow{trainavg}-2*\thisrow{trainstd}] {LeNet_MN.dat};
\addplot [fill=blue!10] fill between[of=upper and lower];

        \addplot [name path=upper,draw=none] table[x=E,y expr=\thisrow{testavg}+2*\thisrow{teststd}] {LeNet_MN.dat};
\addplot [name path=lower,draw=none] table[x=E,y expr=\thisrow{testavg}-2*\thisrow{teststd}] {LeNet_MN.dat};
\addplot [fill=red!10] fill between[of=upper and lower];
    
    \addplot [name path=upper,draw=none] table[x=E,y expr=\thisrow{fastavg}+2*\thisrow{faststd}] {LeNet_MN.dat};
\addplot [name path=lower,draw=none] table[x=E,y expr=\thisrow{fastavg}-2*\thisrow{faststd}] {LeNet_MN.dat};
\addplot [fill=black!10] fill between[of=upper and lower];

    \addplot [name path=upper,draw=none] table[x=E,y expr=\thisrow{slowavg}+2*\thisrow{slowstd}] {LeNet_MN.dat};
\addplot [name path=lower,draw=none] table[x=E,y expr=\thisrow{slowavg}-2*\thisrow{slowstd}] {LeNet_MN.dat};
\addplot [fill=brown!10] fill between[of=upper and lower];
    \end{axis}
    \end{tikzpicture}
    }%
    \caption{LeNet-5, no momentum}
    \label{fig:lenet5Mnist-E}
\end{subfigure}
\renewcommand{\thesubfigure}{c}
\begin{subfigure}{.33\textwidth}
    \centering
    \resizebox{.99\textwidth}{!}{%
    \begin{tikzpicture}
    \begin{axis}[ymin=0, ymax=1.0, legend style={at={(0.27 ,0.95)},anchor=north},xmin=0,xmax=200,
         xlabel=Training epochs,
         ylabel=Classification error]
    \addplot[brown,dashed,mark=square*] table [y=slowavg, x=E]{LeNet_MN_E_low_thresh.dat};
    \addplot[black,dashed,mark=diamond*] table [y=fastavg, x=E]{LeNet_MN_E_low_thresh.dat};
    \addplot[red,solid,mark=otimes*] table [y=testavg, x=E]{LeNet_MN_E_low_thresh.dat};
    \addplot[blue,solid,mark=triangle*] table [y=trainavg, x=E]{LeNet_MN_E_low_thresh.dat};

    \addplot [name path=upper,draw=none] table[x=E,y expr=\thisrow{trainavg}+2*\thisrow{trainstd}] {LeNet_MN_E_low_thresh.dat};
\addplot [name path=lower,draw=none] table[x=E,y expr=\thisrow{trainavg}-2*\thisrow{trainstd}] {LeNet_MN_E_low_thresh.dat};
\addplot [fill=blue!10] fill between[of=upper and lower];
    
        \addplot [name path=upper,draw=none] table[x=E,y expr=\thisrow{testavg}+2*\thisrow{teststd}] {LeNet_MN_E_low_thresh.dat};
\addplot [name path=lower,draw=none] table[x=E,y expr=\thisrow{testavg}-2*\thisrow{teststd}] {LeNet_MN_E_low_thresh.dat};
\addplot [fill=red!10] fill between[of=upper and lower];
    
    \addplot [name path=upper,draw=none] table[x=E,y expr=\thisrow{fastavg}+2*\thisrow{faststd}] {LeNet_MN_E_low_thresh.dat};
\addplot [name path=lower,draw=none] table[x=E,y expr=\thisrow{fastavg}-2*\thisrow{faststd}] {LeNet_MN_E_low_thresh.dat};
\addplot [fill=black!10] fill between[of=upper and lower];

    \addplot [name path=upper,draw=none] table[x=E,y expr=\thisrow{slowavg}+2*\thisrow{slowstd}] {LeNet_MN_E_low_thresh.dat};
\addplot [name path=lower,draw=none] table[x=E,y expr=\thisrow{slowavg}-2*\thisrow{slowstd}] {LeNet_MN_E_low_thresh.dat};
\addplot [fill=brown!10] fill between[of=upper and lower];
    \end{axis}
    \end{tikzpicture}
    }%
    \caption{LeNet-5, momentum, low threshold}
    \label{fig:epochs_low}
\end{subfigure}
\renewcommand{\thesubfigure}{e}
\begin{subfigure}{.33\textwidth}
    \centering
    \resizebox{.99\textwidth}{!}{%
    \begin{tikzpicture}
    \begin{axis}[ymin=0, ymax=1.0, legend style={at={(0.60,0.90)},anchor=north},xmin=5000,xmax=30000,
         xlabel=Training set size $n$,
         ylabel=Classification error]
    \addplot[brown,dashed,mark=square*] table [y=slowavg, x=E]{LeNet_MN_n.dat};
    \addlegendentry{Slow-rate bound}
    \addplot[black,dashed,mark=diamond*] table [y=fastavg, x=E]{LeNet_MN_n.dat};
    \addlegendentry{Fast-rate bound}
    \addplot[red,solid,mark=otimes*] table [y=testavg, x=E]{LeNet_MN_n.dat};
    \addlegendentry{Test loss}
    \addplot[blue,solid,mark=triangle*] table [y=trainavg, x=E]{LeNet_MN_n.dat};
    \addlegendentry{Training loss}
    
    \addplot [name path=upper,draw=none] table[x=E,y expr=\thisrow{trainavg}+2*\thisrow{trainstd}] {LeNet_MN_n.dat};
\addplot [name path=lower,draw=none] table[x=E,y expr=\thisrow{trainavg}-2*\thisrow{trainstd}] {LeNet_MN_n.dat};
\addplot [fill=blue!10] fill between[of=upper and lower];

        \addplot [name path=upper,draw=none] table[x=E,y expr=\thisrow{testavg}+2*\thisrow{teststd}] {LeNet_MN_n.dat};
\addplot [name path=lower,draw=none] table[x=E,y expr=\thisrow{testavg}-2*\thisrow{teststd}] {LeNet_MN_n.dat};
\addplot [fill=red!10] fill between[of=upper and lower];
    
    \addplot [name path=upper,draw=none] table[x=E,y expr=\thisrow{fastavg}+2*\thisrow{faststd}] {LeNet_MN_n.dat};
\addplot [name path=lower,draw=none] table[x=E,y expr=\thisrow{fastavg}-2*\thisrow{faststd}] {LeNet_MN_n.dat};
\addplot [fill=black!10] fill between[of=upper and lower];
    
    \addplot [name path=upper,draw=none] table[x=E,y expr=\thisrow{slowavg}+2*\thisrow{slowstd}] {LeNet_MN_n.dat};
\addplot [name path=lower,draw=none] table[x=E,y expr=\thisrow{slowavg}-2*\thisrow{slowstd}] {LeNet_MN_n.dat};
\addplot [fill=brown!10] fill between[of=upper and lower];
    \end{axis}
    \end{tikzpicture}
    }%
    \caption{LeNet-5}
    \label{fig:lenet5Mnist-n}
\end{subfigure}

\renewcommand{\thesubfigure}{b}
\begin{subfigure}{.33\textwidth}
    \centering
    \resizebox{.99\textwidth}{!}{%
    \begin{tikzpicture}
    \begin{axis}[ymin=0, ymax=1.0, legend style={at={(0.98,0.5)},anchor=east},xmin=0,xmax=100.0,
         xlabel=Training epochs, ylabel=Classification error
         ]
    \addplot[brown,dashed,mark=square*] table [y=slowavg, x=E]{FC_MN.dat};
    \addplot[black,dashed,mark=diamond*] table [y=fastavg, x=E]{FC_MN.dat};
    \addplot[red,solid,mark=otimes*] table [y=testavg, x=E]{FC_MN.dat};
    \addplot[blue,solid,mark=triangle*] table [y=trainavg, x=E]{FC_MN.dat};

    \addplot [name path=upper,draw=none] table[x=E,y expr=\thisrow{trainavg}+2*\thisrow{trainstd}] {FC_MN.dat};
\addplot [name path=lower,draw=none] table[x=E,y expr=\thisrow{trainavg}-2*\thisrow{trainstd}] {FC_MN.dat};
\addplot [fill=blue!10] fill between[of=upper and lower];
    
        \addplot [name path=upper,draw=none] table[x=E,y expr=\thisrow{testavg}+2*\thisrow{teststd}] {FC_MN.dat};
\addplot [name path=lower,draw=none] table[x=E,y expr=\thisrow{testavg}-2*\thisrow{teststd}] {FC_MN.dat};
\addplot [fill=red!10] fill between[of=upper and lower];
    
    \addplot [name path=upper,draw=none] table[x=E,y expr=\thisrow{fastavg}+2*\thisrow{faststd}] {FC_MN.dat};
\addplot [name path=lower,draw=none] table[x=E,y expr=\thisrow{fastavg}-2*\thisrow{faststd}] {FC_MN.dat};
\addplot [fill=black!10] fill between[of=upper and lower];

    \addplot [name path=upper,draw=none] table[x=E,y expr=\thisrow{slowavg}+2*\thisrow{slowstd}] {FC_MN.dat};
\addplot [name path=lower,draw=none] table[x=E,y expr=\thisrow{slowavg}-2*\thisrow{slowstd}] {FC_MN.dat};
\addplot [fill=brown!10] fill between[of=upper and lower];
    \end{axis}
    \end{tikzpicture}
    }%
    \caption{$600^2$ FCNN, no momentum}
    \label{fig:FCNNMnist-e}
\end{subfigure}
\renewcommand{\thesubfigure}{d}
\begin{subfigure}{.33\textwidth}
    \centering
    \resizebox{.99\textwidth}{!}{%
    \begin{tikzpicture}
    \begin{axis}[ymin=0, ymax=1.0, legend style={at={(0.60,1.0)},anchor=north},xmin=0,xmax=200,
         xlabel=Training epochs,
         ylabel=Classification error]
    \addplot[brown,dashed,mark=square*] table [y=slowavg, x=E]{LeNet_MN_E_high_thresh.dat};
    \addplot[black,dashed,mark=diamond*] table [y=fastavg, x=E]{LeNet_MN_E_high_thresh.dat};
    \addplot[red,solid,mark=otimes*] table [y=testavg, x=E]{LeNet_MN_E_high_thresh.dat};
    \addplot[blue,solid,mark=triangle*] table [y=trainavg, x=E]{LeNet_MN_E_high_thresh.dat};
    
    \addplot [name path=upper,draw=none] table[x=E,y expr=\thisrow{trainavg}+2*\thisrow{trainstd}] {LeNet_MN_E_high_thresh.dat};
\addplot [name path=lower,draw=none] table[x=E,y expr=\thisrow{trainavg}-2*\thisrow{trainstd}] {LeNet_MN_E_high_thresh.dat};
\addplot [fill=blue!10] fill between[of=upper and lower];
    
        \addplot [name path=upper,draw=none] table[x=E,y expr=\thisrow{testavg}+2*\thisrow{teststd}] {LeNet_MN_E_high_thresh.dat};
\addplot [name path=lower,draw=none] table[x=E,y expr=\thisrow{testavg}-2*\thisrow{teststd}] {LeNet_MN_E_high_thresh.dat};
\addplot [fill=red!10] fill between[of=upper and lower];
    
    \addplot [name path=upper,draw=none] table[x=E,y expr=\thisrow{fastavg}+2*\thisrow{faststd}] {LeNet_MN_E_high_thresh.dat};
\addplot [name path=lower,draw=none] table[x=E,y expr=\thisrow{fastavg}-2*\thisrow{faststd}] {LeNet_MN_E_high_thresh.dat};
\addplot [fill=black!10] fill between[of=upper and lower];

    \addplot [name path=upper,draw=none] table[x=E,y expr=\thisrow{slowavg}+2*\thisrow{slowstd}] {LeNet_MN_E_high_thresh.dat};
\addplot [name path=lower,draw=none] table[x=E,y expr=\thisrow{slowavg}-2*\thisrow{slowstd}] {LeNet_MN_E_high_thresh.dat};
\addplot [fill=brown!10] fill between[of=upper and lower];
    \end{axis}
    \end{tikzpicture}
    }%
    \caption{LeNet-5, momentum, high threshold}
    \label{fig:epochs_high}
\end{subfigure}
\renewcommand{\thesubfigure}{f}
\begin{subfigure}{.33\textwidth}
    \centering
    \resizebox{.99\textwidth}{!}{%
    \begin{tikzpicture}
    \begin{axis}[ymin=0, ymax=1.0, legend style={at={(0.60,0.90)},anchor=north},xmin=5000,xmax=30000,
         xlabel=Training set size $n$,
         ylabel=Classification error]
    \addplot[brown,dashed,mark=square*] table [y=slowavg, x=E]{FC_MN_n.dat};
    \addplot[black,dashed,mark=diamond*] table [y=fastavg, x=E]{FC_MN_n.dat};
    \addplot[red,solid,mark=otimes*] table [y=testavg, x=E]{FC_MN_n.dat};
    \addplot[blue,solid,mark=triangle*] table [y=trainavg, x=E]{FC_MN_n.dat};
    
    \addplot [name path=upper,draw=none] table[x=E,y expr=\thisrow{trainavg}+2*\thisrow{trainstd}] {FC_MN_n.dat};
\addplot [name path=lower,draw=none] table[x=E,y expr=\thisrow{trainavg}-2*\thisrow{trainstd}] {FC_MN_n.dat};
\addplot [fill=blue!10] fill between[of=upper and lower];
    
        \addplot [name path=upper,draw=none] table[x=E,y expr=\thisrow{testavg}+2*\thisrow{teststd}] {FC_MN_n.dat};
\addplot [name path=lower,draw=none] table[x=E,y expr=\thisrow{testavg}-2*\thisrow{teststd}] {FC_MN_n.dat};
\addplot [fill=red!10] fill between[of=upper and lower];
    
    \addplot [name path=upper,draw=none] table[x=E,y expr=\thisrow{fastavg}+2*\thisrow{faststd}] {FC_MN_n.dat};
\addplot [name path=lower,draw=none] table[x=E,y expr=\thisrow{fastavg}-2*\thisrow{faststd}] {FC_MN_n.dat};
\addplot [fill=black!10] fill between[of=upper and lower];

    \addplot [name path=upper,draw=none] table[x=E,y expr=\thisrow{slowavg}+2*\thisrow{slowstd}] {FC_MN_n.dat};
\addplot [name path=lower,draw=none] table[x=E,y expr=\thisrow{slowavg}-2*\thisrow{slowstd}] {FC_MN_n.dat};
\addplot [fill=brown!10] fill between[of=upper and lower];
    \end{axis}
    \end{tikzpicture}
    }%
    \caption{$600^2$ FCNN}
    \label{fig:FCNNMnist-n}
\end{subfigure}

\caption{The estimated training losses and test losses as well as the slow-rate~\eqref{eq:hellstrom_slow_pacb} and fast-rate~\eqref{eq:pacb_bound} PAC-Bayesian bounds on the test loss for two NNs trained on MNIST. The shaded regions correspond to two standard deviations. In (a)--(b), we perform training using SGD without momentum with a decaying learning rate. In (c)--(d), we use SGD with momentum and a fixed learning rate. In (e)--(f), for each value of $n$, we use SGD with momentum until a target training loss is reached. Further details on the experimental setup are given in Appendix~\ref{app:experiment_details}.
    }
    \label{fig:plots_part_1}
\end{figure}

With these choices, we can explicitly evaluate the bounds. 
For each setting, we perform simulations over $10$ instances of $\subsetchoice$. 
Our results are obtained by setting~$\delta\approx 0.001$ as the confidence parameter. 
However, since the bounds are optimized over the choice of $\sigma_2$, we need to use a union bound argument~\cite{dziugaite-20,Dziugaite2017} to guarantee that the final slow-rate and fast-rate bounds hold for all of these candidates simultaneously. 
As a consequence, the presented bounds hold with probability at least~$95\%$. 
The test loss and training loss are computed empirically by averaging the performance of~$5$ NNs whose weights are sampled from~$\normal(W\mid \boldsymbol{\mu}_1,\sigma^2_1\matI_d)$.

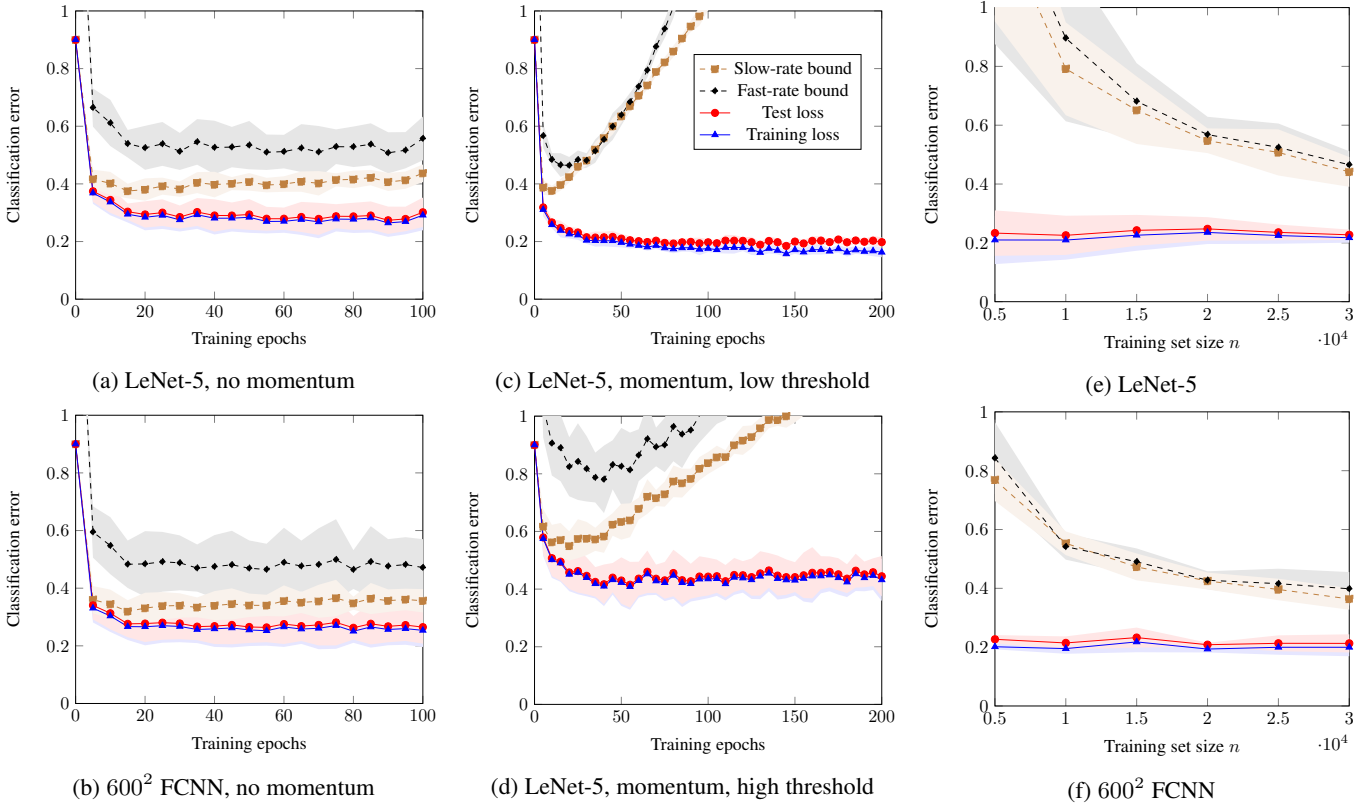
\begin{figure}

\renewcommand{\thesubfigure}{a}
\begin{subfigure}{.33\textwidth}
    \centering
    \resizebox{.99\textwidth}{!}{%
    \begin{tikzpicture}
    \begin{axis}[ymin=0, ymax=1.0, legend style={at={(0.01,0.42)},anchor=west},xmin=0,xmax=100.0,
         xlabel=Training epochs,
         ylabel=Classification error]
    \addplot[brown,dashed,mark=square*] table [y=slowavg, x=E]{LeNet_FMN.dat};
    \addplot[black,dashed,mark=diamond*] table [y=fastavg, x=E]{LeNet_FMN.dat};
    \addplot[red,solid,mark=otimes*] table [y=testavg, x=E]{LeNet_FMN.dat};
    \addplot[blue,solid,mark=triangle*] table [y=trainavg, x=E]{LeNet_FMN.dat};

    \addplot [name path=upper,draw=none] table[x=E,y expr=\thisrow{trainavg}+2*\thisrow{trainstd}] {LeNet_FMN.dat};
\addplot [name path=lower,draw=none] table[x=E,y expr=\thisrow{trainavg}-2*\thisrow{trainstd}] {LeNet_FMN.dat};
\addplot [fill=blue!10] fill between[of=upper and lower];
    
        \addplot [name path=upper,draw=none] table[x=E,y expr=\thisrow{testavg}+2*\thisrow{teststd}] {LeNet_FMN.dat};
\addplot [name path=lower,draw=none] table[x=E,y expr=\thisrow{testavg}-2*\thisrow{teststd}] {LeNet_FMN.dat};
\addplot [fill=red!10] fill between[of=upper and lower];
    
    \addplot [name path=upper,draw=none] table[x=E,y expr=\thisrow{fastavg}+2*\thisrow{faststd}] {LeNet_FMN.dat};
\addplot [name path=lower,draw=none] table[x=E,y expr=\thisrow{fastavg}-2*\thisrow{faststd}] {LeNet_FMN.dat};
\addplot [fill=black!10] fill between[of=upper and lower];

    \addplot [name path=upper,draw=none] table[x=E,y expr=\thisrow{slowavg}+2*\thisrow{slowstd}] {LeNet_FMN.dat};
\addplot [name path=lower,draw=none] table[x=E,y expr=\thisrow{slowavg}-2*\thisrow{slowstd}] {LeNet_FMN.dat};
\addplot [fill=brown!10] fill between[of=upper and lower];
    \end{axis}
    \end{tikzpicture}
    }%
    \caption{LeNet-5, no momentum}
\end{subfigure}
\renewcommand{\thesubfigure}{c}
\begin{subfigure}{.33\textwidth}
    \centering
    \resizebox{.99\textwidth}{!}{%
    \begin{tikzpicture}
    \begin{axis}[ymin=0, ymax=1.000, legend style={at={(0.70 ,0.85)},anchor=north},xmin=0,xmax=200,
         xlabel=Training epochs,
         ylabel=Classification error]
    \addplot[brown,dashed,mark=square*] table [y=slowavg, x=E]{LeNet_FMN_E_low_thresh.dat};
    \addlegendentry{Slow-rate bound}
    \addplot[black,dashed,mark=diamond*] table [y=fastavg, x=E]{LeNet_FMN_E_low_thresh.dat};
    \addlegendentry{Fast-rate bound}
    \addplot[red,solid,mark=otimes*] table [y=testavg, x=E]{LeNet_FMN_E_low_thresh.dat};
    \addlegendentry{Test loss}
    \addplot[blue,solid,mark=triangle*] table [y=trainavg, x=E]{LeNet_FMN_E_low_thresh.dat};
    \addlegendentry{Training loss}
    
    \addplot [name path=upper,draw=none] table[x=E,y expr=\thisrow{trainavg}+2*\thisrow{trainstd}] {LeNet_FMN_E_low_thresh.dat};
\addplot [name path=lower,draw=none] table[x=E,y expr=\thisrow{trainavg}-2*\thisrow{trainstd}] {LeNet_FMN_E_low_thresh.dat};
\addplot [fill=blue!10] fill between[of=upper and lower];
    
        \addplot [name path=upper,draw=none] table[x=E,y expr=\thisrow{testavg}+2*\thisrow{teststd}] {LeNet_FMN_E_low_thresh.dat};
\addplot [name path=lower,draw=none] table[x=E,y expr=\thisrow{testavg}-2*\thisrow{teststd}] {LeNet_FMN_E_low_thresh.dat};
\addplot [fill=red!10] fill between[of=upper and lower];
    
    \addplot [name path=upper,draw=none] table[x=E,y expr=\thisrow{fastavg}+2*\thisrow{faststd}] {LeNet_FMN_E_low_thresh.dat};
\addplot [name path=lower,draw=none] table[x=E,y expr=\thisrow{fastavg}-2*\thisrow{faststd}] {LeNet_FMN_E_low_thresh.dat};
\addplot [fill=black!10] fill between[of=upper and lower];

    \addplot [name path=upper,draw=none] table[x=E,y expr=\thisrow{slowavg}+2*\thisrow{slowstd}] {LeNet_FMN_E_low_thresh.dat};
\addplot [name path=lower,draw=none] table[x=E,y expr=\thisrow{slowavg}-2*\thisrow{slowstd}] {LeNet_FMN_E_low_thresh.dat};
\addplot [fill=brown!10] fill between[of=upper and lower];
    \end{axis}
    \end{tikzpicture}
    }%
    \caption{LeNet-5, momentum, low threshold}
\end{subfigure}
\renewcommand{\thesubfigure}{e}
\begin{subfigure}{.33\textwidth}
    \centering
    \resizebox{.99\textwidth}{!}{%
    \begin{tikzpicture}
    \begin{axis}[ymin=0, ymax=1.0, legend style={at={(0.01,0.42)},anchor=west},xmin=5000,xmax=30000,
         xlabel=Training set size $n$,
         ylabel=Classification error]
    \addplot[brown,dashed,mark=square*] table [y=slowavg, x=E]{LeNet_FMN_n.dat};
    \addplot[black,dashed,mark=diamond*] table [y=fastavg, x=E]{LeNet_FMN_n.dat};
    \addplot[red,solid,mark=otimes*] table [y=testavg, x=E]{LeNet_FMN_n.dat};
    \addplot[blue,solid,mark=triangle*] table [y=trainavg, x=E]{LeNet_FMN_n.dat};

    \addplot [name path=upper,draw=none] table[x=E,y expr=\thisrow{trainavg}+2*\thisrow{trainstd}] {LeNet_FMN_n.dat};
\addplot [name path=lower,draw=none] table[x=E,y expr=\thisrow{trainavg}-2*\thisrow{trainstd}] {LeNet_FMN_n.dat};
\addplot [fill=blue!10] fill between[of=upper and lower];
    
        \addplot [name path=upper,draw=none] table[x=E,y expr=\thisrow{testavg}+2*\thisrow{teststd}] {LeNet_FMN_n.dat};
\addplot [name path=lower,draw=none] table[x=E,y expr=\thisrow{testavg}-2*\thisrow{teststd}] {LeNet_FMN_n.dat};
\addplot [fill=red!10] fill between[of=upper and lower];
    
    \addplot [name path=upper,draw=none] table[x=E,y expr=\thisrow{fastavg}+2*\thisrow{faststd}] {LeNet_FMN_n.dat};
\addplot [name path=lower,draw=none] table[x=E,y expr=\thisrow{fastavg}-2*\thisrow{faststd}] {LeNet_FMN_n.dat};
\addplot [fill=black!10] fill between[of=upper and lower];

    \addplot [name path=upper,draw=none] table[x=E,y expr=\thisrow{slowavg}+2*\thisrow{slowstd}] {LeNet_FMN_n.dat};
\addplot [name path=lower,draw=none] table[x=E,y expr=\thisrow{slowavg}-2*\thisrow{slowstd}] {LeNet_FMN_n.dat};
\addplot [fill=brown!10] fill between[of=upper and lower];
    \end{axis}
    \end{tikzpicture}
    }%
    \caption{LeNet-5}
\end{subfigure}

\renewcommand{\thesubfigure}{b}
\begin{subfigure}{.33\textwidth}
    \centering
    \resizebox{.99\textwidth}{!}{%
    \begin{tikzpicture}
    \begin{axis}[ymin=0, ymax=1.0, legend style={at={(0.98,0.4)},anchor=east},xmin=0,xmax=100.0,
         xlabel=Training epochs, ylabel=Classification error
         ]
    \addplot[brown,dashed,mark=square*] table [y=slowavg, x=E]{FC_FMN.dat};
    \addplot[black,dashed,mark=diamond*] table [y=fastavg, x=E]{FC_FMN.dat};
    \addplot[red,solid,mark=otimes*] table [y=testavg, x=E]{FC_FMN.dat};
    \addplot[blue,solid,mark=triangle*] table [y=trainavg, x=E]{FC_FMN.dat};

    \addplot [name path=upper,draw=none] table[x=E,y expr=\thisrow{trainavg}+2*\thisrow{trainstd}] {FC_FMN.dat};
\addplot [name path=lower,draw=none] table[x=E,y expr=\thisrow{trainavg}-2*\thisrow{trainstd}] {FC_FMN.dat};
\addplot [fill=blue!10] fill between[of=upper and lower];

        \addplot [name path=upper,draw=none] table[x=E,y expr=\thisrow{testavg}+2*\thisrow{teststd}] {FC_FMN.dat};
\addplot [name path=lower,draw=none] table[x=E,y expr=\thisrow{testavg}-2*\thisrow{teststd}] {FC_FMN.dat};
\addplot [fill=red!10] fill between[of=upper and lower];
    
    \addplot [name path=upper,draw=none] table[x=E,y expr=\thisrow{fastavg}+2*\thisrow{faststd}] {FC_FMN.dat};
\addplot [name path=lower,draw=none] table[x=E,y expr=\thisrow{fastavg}-2*\thisrow{faststd}] {FC_FMN.dat};
\addplot [fill=black!10] fill between[of=upper and lower];

    \addplot [name path=upper,draw=none] table[x=E,y expr=\thisrow{slowavg}+2*\thisrow{slowstd}] {FC_FMN.dat};
\addplot [name path=lower,draw=none] table[x=E,y expr=\thisrow{slowavg}-2*\thisrow{slowstd}] {FC_FMN.dat};
\addplot [fill=brown!10] fill between[of=upper and lower];
    \end{axis}
    \end{tikzpicture}
    }%
    \caption{$600^2$ FCNN, no momentum}
\end{subfigure}
\renewcommand{\thesubfigure}{d}
\begin{subfigure}{.33\textwidth}
    \centering
    \resizebox{.99\textwidth}{!}{%
    \begin{tikzpicture}
    \begin{axis}[ymin=0, ymax=1.000, legend style={at={(0.60,1.0)},anchor=north},xmin=0,xmax=200,
         xlabel=Training epochs,
         ylabel=Classification error]
    \addplot[brown,dashed,mark=square*] table [y=slowavg, x=E]{LeNet_FMN_E_high_thresh.dat};
    \addplot[black,dashed,mark=diamond*] table [y=fastavg, x=E]{LeNet_FMN_E_high_thresh.dat};
    \addplot[red,solid,mark=otimes*] table [y=testavg, x=E]{LeNet_FMN_E_high_thresh.dat};
    \addplot[blue,solid,mark=triangle*] table [y=trainavg, x=E]{LeNet_FMN_E_high_thresh.dat};
    
    \addplot [name path=upper,draw=none] table[x=E,y expr=\thisrow{trainavg}+2*\thisrow{trainstd}] {LeNet_FMN_E_high_thresh.dat};
\addplot [name path=lower,draw=none] table[x=E,y expr=\thisrow{trainavg}-2*\thisrow{trainstd}] {LeNet_FMN_E_high_thresh.dat};
\addplot [fill=blue!10] fill between[of=upper and lower];
    
        \addplot [name path=upper,draw=none] table[x=E,y expr=\thisrow{testavg}+2*\thisrow{teststd}] {LeNet_FMN_E_high_thresh.dat};
\addplot [name path=lower,draw=none] table[x=E,y expr=\thisrow{testavg}-2*\thisrow{teststd}] {LeNet_FMN_E_high_thresh.dat};
\addplot [fill=red!10] fill between[of=upper and lower];
    
    \addplot [name path=upper,draw=none] table[x=E,y expr=\thisrow{fastavg}+2*\thisrow{faststd}] {LeNet_FMN_E_high_thresh.dat};
\addplot [name path=lower,draw=none] table[x=E,y expr=\thisrow{fastavg}-2*\thisrow{faststd}] {LeNet_FMN_E_high_thresh.dat};
\addplot [fill=black!10] fill between[of=upper and lower];

    \addplot [name path=upper,draw=none] table[x=E,y expr=\thisrow{slowavg}+2*\thisrow{slowstd}] {LeNet_FMN_E_high_thresh.dat};
\addplot [name path=lower,draw=none] table[x=E,y expr=\thisrow{slowavg}-2*\thisrow{slowstd}] {LeNet_FMN_E_high_thresh.dat};
\addplot [fill=brown!10] fill between[of=upper and lower];
    \end{axis}
    \end{tikzpicture}
    }%
    \caption{LeNet-5, momentum, high threshold}
\end{subfigure}
\renewcommand{\thesubfigure}{f}
\begin{subfigure}{.33\textwidth}
    \centering
    \resizebox{.99\textwidth}{!}{%
    \begin{tikzpicture}
    \begin{axis}[ymin=0, ymax=1.0, legend style={at={(0.98,0.4)},anchor=east},xmin=5000,xmax=30000,
         xlabel=Training set size $n$, ylabel=Classification error
         ]
    \addplot[brown,dashed,mark=square*] table [y=slowavg, x=E]{FC_FMN_n.dat};
    \addplot[black,dashed,mark=diamond*] table [y=fastavg, x=E]{FC_FMN_n.dat};
    \addplot[red,solid,mark=otimes*] table [y=testavg, x=E]{FC_FMN_n.dat};
    \addplot[blue,solid,mark=triangle*] table [y=trainavg, x=E]{FC_FMN_n.dat};

    \addplot [name path=upper,draw=none] table[x=E,y expr=\thisrow{trainavg}+2*\thisrow{trainstd}] {FC_FMN_n.dat};
\addplot [name path=lower,draw=none] table[x=E,y expr=\thisrow{trainavg}-2*\thisrow{trainstd}] {FC_FMN_n.dat};
\addplot [fill=blue!10] fill between[of=upper and lower];

        \addplot [name path=upper,draw=none] table[x=E,y expr=\thisrow{testavg}+2*\thisrow{teststd}] {FC_FMN_n.dat};
\addplot [name path=lower,draw=none] table[x=E,y expr=\thisrow{testavg}-2*\thisrow{teststd}] {FC_FMN_n.dat};
\addplot [fill=red!10] fill between[of=upper and lower];
    
    \addplot [name path=upper,draw=none] table[x=E,y expr=\thisrow{fastavg}+2*\thisrow{faststd}] {FC_FMN_n.dat};
\addplot [name path=lower,draw=none] table[x=E,y expr=\thisrow{fastavg}-2*\thisrow{faststd}] {FC_FMN_n.dat};
\addplot [fill=black!10] fill between[of=upper and lower];

    \addplot [name path=upper,draw=none] table[x=E,y expr=\thisrow{slowavg}+2*\thisrow{slowstd}] {FC_FMN_n.dat};
\addplot [name path=lower,draw=none] table[x=E,y expr=\thisrow{slowavg}-2*\thisrow{slowstd}] {FC_FMN_n.dat};
\addplot [fill=brown!10] fill between[of=upper and lower];
    \end{axis}
    \end{tikzpicture}
    }%
    \caption{$600^2$ FCNN}
\end{subfigure}
\caption{The estimated training losses and test losses as well as the slow-rate~\eqref{eq:hellstrom_slow_pacb} and fast-rate~\eqref{eq:pacb_bound} PAC-Bayesian bounds on the test loss for two NNs trained on Fashion-MNIST. The shaded regions correspond to two standard deviations. In (a)--(b), we perform training using SGD without momentum with a decaying learning rate. In (c)--(d), we use SGD with momentum and a fixed learning rate. In (e)--(f), for each value of $n$, we use SGD with momentum until a target training loss is reached. Further details on the experimental setup are given in Appendix~\ref{app:experiment_details}.
    }
    \label{fig:plots_part_2}
\end{figure}

In Figure~\ref{fig:plots_part_1}, we plot the slow-rate and fast-rate bounds, as well as the estimated training and test losses, for networks trained on the MNIST data set. 
In Figures~\ref{fig:plots_part_1}a--d, we plot these metrics as a function of the number of training epochs. 
In Figures~\ref{fig:plots_part_1}a--b, we train the NNs using SGD with a decaying learning rate without momentum, whereas in Figures~\ref{fig:plots_part_1}c--d, we use SGD with momentum and a fixed learning rate. 
In Figures~\ref{fig:plots_part_1}e--f, we restrict $\supersample$ to only be a subset of the $6\cdot 10^4$ available training samples from MNIST, and train the NNs using momentum until they reach a target training loss (0.05 for MNIST and 0.15 for Fashion-MNIST). 
We then present the results as a function of the size $n$ of the training set $\trainingdata(\subsetchoice)$. 
In Figure~\ref{fig:plots_part_2}, we present the corresponding results for the Fashion-MNIST data set. 
In Table~\ref{tab:randomized_labels}, we replace a portion of the data labels with a randomly chosen label, and study how the proportion of corrupt data affects our bounds. 
In order to make training with randomized labels more efficient, we consider a binarized version of MNIST where the digits $0,\dots,4$ are combined into one class and the digits $5,\dots,9$ into another. 
Detailed descriptions of the architectures and training procedures are given in Appendix~\ref{app:experiment_details}.

In Figure~\ref{fig:plots_part_1}, for the MNIST data set, we see that the fast-rate bound tends to be tighter than its slow-rate counterpart. For the more challenging Fashion-MNIST data set in Figure~\ref{fig:plots_part_2}, the slow-rate bound is tighter. This is due to the fact that high training losses and large information measures penalize the fast-rate bound more due to its larger constant factors.

The quantitative values of our bounds in Figures~\ref{fig:plots_part_1}a--b and ~\ref{fig:plots_part_2}a--b are in line with previously reported results for a similar setup~\cite[Fig.~4]{dziugaite-20}. 
The minimum test-loss bounds (averaged over~$50$ runs) for MNIST reported in~\cite[Fig.~4]{dziugaite-20} are approximately $0.13$ for LeNet-5 and $0.18$ for the $600^2$ FCNN.
These values are similar to our best bounds, which are~$0.15$ for LeNet-5 and~$0.19$ for the $600^2$ FCNN. 
For LeNet-5 trained on Fashion-MNIST, our tightest bound on the test loss is $0.35$, whereas the corresponding one in~\cite[Fig.~4]{dziugaite-20} is approximately $0.36$.
Taking error bars into account, our bounds are not clearly distinguishable from those reported in~\cite[Fig.~4]{dziugaite-20}.
It is important to mention that significantly tighter bounds are reported in~\cite[Fig.~5]{dziugaite-20} for the case in which the PAC-Bayesian bound considered therein is used as a regularizer during the training process. 
Such a direct optimization of the bound does not appear to be feasible for the random-subset setting considered in this paper.

\begin{table}
\caption{The estimated training losses, test losses, and the corresponding slow-rate~\eqref{eq:hellstrom_slow_pacb} and fast-rate~\eqref{eq:pacb_bound} PAC-Bayesian bounds on the test loss for LeNet-5 trained on binarized MNIST with partially corrupted labels. }
\label{tab:randomized_labels}
\centering
\begin{tabular}{lccccc}
\toprule Randomized labels & $25\%$ & $50\%$ & $75\%$ & $100\%$ \\
\midrule
Training loss  & $0.106$ & $0.088$ & $0.090$ & $0.081$   \\
Test loss & $0.216$ & $0.364$ & $0.461$ & $0.494$ \\
Slow-rate bound & $5.561$ & $9.811$ & $10.45$ & $11.67$ \\
Fast-rate bound & $44.52$ & $141.0$ & $160.1$ & $200.4$ \\ \bottomrule
\end{tabular}
\end{table}

Next, we discuss the results presented in Figures~\ref{fig:plots_part_1}c--d and \ref{fig:plots_part_2}c--d, where we consider SGD with momentum.
While our bounds become tighter in the initial phase of training, they lose tightness as training progresses and smaller training errors (on the order of $0.001$) are reached for the deterministic NNs. 
This is similar to what is noted by~\cite[p.~12]{dziugaite-20}.
Specifically, when the underlying deterministic NN therein is trained to achieve very low errors (or equivalently, is trained for many epochs), the PAC-Bayesian bound they consider becomes loose, and the corresponding stochastic NN has a significantly higher test error than the underlying deterministic NN. 
The difference in behavior of our bounds in Figure~\ref{fig:plots_part_1}c and Figure~\ref{fig:plots_part_1}d illustrates the role played by the variances~$\sigma_1$ and~$\sigma_2$.  In Figure~\ref{fig:plots_part_1}c, we set the threshold used to determine $\sigma_1$ and $\sigma_2$ to $0.05$, which leads to small values for~$\sigma_1$ and~$\sigma_2$. 
In Figure~\ref{fig:plots_part_1}d, we use a threshold of $0.15$ instead, which allows for larger variances. 
The results illustrate the intuitive fact that larger variances yield better test-loss bounds at the cost of a higher true test error. A similar observation can be made for Fashion-MNIST in Figures~\ref{fig:plots_part_2}c--d. For Figure~\ref{fig:plots_part_2}c, we use a threshold of $0.10$, while for Figure~\ref{fig:plots_part_2}d, we set it to $0.45$.

While the decay in $n$ seen in the bounds in Figures~\ref{fig:plots_part_1}e--f and~\ref{fig:plots_part_2}e--f shows that the information measures therein are sublinear in $n$, we note that the similarity in behavior for the fast- and slow-rate bounds indicates that the growth is almost linear. Thus, for the particular priors and posteriors that we study, the sublinearity appears to be mild. Furthermore, the bounds are vacuous for low values of $n$, despite nearly perfect generalization being achieved in practice. 

As shown in Table~\ref{tab:randomized_labels}, our bounds become vacuous when randomized labels are used. 
The fast-rate bound is significantly worse than its slow-rate counterpart, which is to be expected: when the prior and posterior are selected using randomized labels, a larger discrepancy between them arises. 
This increases the value of the KL divergence in \eqref{eq:hellstrom_slow_pacb} and \eqref{eq:pacb_bound}, which, as previously discussed, penalizes the fast-rate bound more. 
We note, though, that the qualitative behavior of the bounds is in agreement with the empirically evaluated test error: an increased proportion of randomized labels, and thus an increased test error, is accompanied by an increase in the values of our bounds. Furthermore, 
the slow-rate bound consistently overestimates the test error by a factor of approximately~$25$.

\section{Conclusion}
We have studied information-theoretic bounds on the test loss in the random-subset setting, in which the posterior and the training loss depend on a randomly selected subset of the available data set, and the prior is allowed to depend on the entire data set. 
In particular, we derived new fast-rate bounds for the PAC-Bayesian and single-draw settings. 
Provided that the information measures appearing in the bounds scale sublinearly with~$n$, these fast-rate bounds have a better asymptotic dependence on~$n$ than the slow-rate PAC-Bayesian and single-draw bounds previously reported in~\cite{hellstrom-20b}, at the price of larger multiplicative constants. 

Through numerical experiments, we show that our novel fast-rate PAC-Bayesian bound, as well as its slow-rate counterpart, result in test-loss bounds for some overparameterized NNs trained through SGD that are in line with previously reported bounds in the literature~\cite{dziugaite-20}.
Furthermore, the single-draw counterparts of these bounds, which are as tight as the PAC-Bayesian bounds, are applicable also to deterministic NNs trained through SGD and with Gaussian noise added to the final weights. 
On the negative side, as illustrated in Figures~\ref{fig:plots_part_1}c--d and~\ref{fig:plots_part_2}c--d, the bounds turn out to be loose when applied to NNs trained to achieve very small training errors. 
Moreover, Figures~\ref{fig:plots_part_1}e--f and~\ref{fig:plots_part_2}e--f reveal that the bounds overestimate the number of training samples needed to guarantee generalization, while Table~\ref{tab:randomized_labels} shows that they become vacuous when randomized labels are introduced.

Still, the results demonstrate the value of the random-subset approach in studying the generalization capabilities of NNs, and show that fast-rate versions of the available information-theoretic bounds can be beneficial in this setting. 
In particular, the random-subset setting provides a natural way to select data-dependent priors, namely by marginalizing the learning algorithm~$P_{W\vert\supersample\!\subsetchoice}$ over~$\subsetchoice$, either exactly or approximately. Such data-dependent priors are a key element in obtaining tight information-theoretic generalization bounds~\cite{dziugaite-20}.

\bibliographystyle{IEEEtran}
\bibliography{reference}

\begin{thebibliography}{10}
\providecommand{\url}[1]{#1}
\csname url@samestyle\endcsname
\providecommand{\newblock}{\relax}
\providecommand{\bibinfo}[2]{#2}
\providecommand{\BIBentrySTDinterwordspacing}{\spaceskip=0pt\relax}
\providecommand{\BIBentryALTinterwordstretchfactor}{4}
\providecommand{\BIBentryALTinterwordspacing}{\spaceskip=\fontdimen2\font plus
\BIBentryALTinterwordstretchfactor\fontdimen3\font minus
  \fontdimen4\font\relax}
\providecommand{\BIBforeignlanguage}[2]{{%
\expandafter\ifx\csname l@#1\endcsname\relax
\typeout{** WARNING: IEEEtran.bst: No hyphenation pattern has been}%
\typeout{** loaded for the language `#1'. Using the pattern for}%
\typeout{** the default language instead.}%
\else
\language=\csname l@#1\endcsname
\fi
#2}}
\providecommand{\BIBdecl}{\relax}
\BIBdecl

\bibitem{mcallester98-07a}
D.~McAllester, ``Some {PAC-B}ayesian theorems,'' in \emph{Proc. Conf. Learn.
  Theory (COLT)}, Madison, WI, July 1998.

\bibitem{catoni07-a}
O.~Catoni, \emph{{PAC}-{{Bayesian Supervised Classification}}: {{The
  Thermodynamics}} of {{Statistical Learning}}}.\hskip 1em plus 0.5em minus
  0.4em\relax IMS Lecture Notes Monogr. Ser., 2007, vol.~56.

\bibitem{guedj19-01a}
\BIBentryALTinterwordspacing
B.~Guedj, ``A {primer} on {PAC}-{Bayesian} {learning},'' \emph{arXiv}, Jan.
  2019. [Online]. Available: \url{http://arxiv.org/abs/1901.05353}
\BIBentrySTDinterwordspacing

\bibitem{russo16-05b}
D.~Russo and J.~Zou, ``Controlling {bias} in {adaptive} {data} {analysis}
  {using} {information} {theory},'' in \emph{Proc. Artif. Intell. Statist.
  (AISTATS)}, Cadiz, Spain, May 2016.

\bibitem{xu17-05a}
A.~Xu and M.~Raginsky, ``Information-theoretic analysis of generalization
  capability of learning algorithms,'' in \emph{Proc. Conf. Neural Inf.
  Process. Syst. (NeurIPS)}, Long Beach, CA, Dec. 2017.

\bibitem{Bu-19-ISIT}
Y.~{Bu}, S.~{Zou}, and V.~V. {Veeravalli}, ``Tightening mutual information
  based bounds on generalization error,'' in \emph{Proc. IEEE Int. Symp. Inf.
  Theory (ISIT)}, Paris, France, July 2019.

\bibitem{Asadi2018}
A.~R. Asadi, E.~Abbe, and S.~Verd{\'{u}}, ``Chaining mutual information and
  tightening generalization bounds,'' in \emph{Proc. Conf. Neural Inf. Process.
  Syst. (NeurIPS)}, Montreal, Canada, Dec. 2018.

\bibitem{Negrea2019}
J.~Negrea, M.~Haghifam, G.~Dziugaite, A.~Khisti, and D.~Roy,
  ``Information-theoretic generalization bounds for {SGLD} via data-dependent
  estimates,'' in \emph{Proc. Conf. Neural Inf. Process. Syst. (NeurIPS)},
  Vancouver, Canada, Dec. 2019.

\bibitem{bassily18-02a}
R.~Bassily, S.~Moran, I.~Nachum, J.~Shafer, and A.~Yehudayoff, ``Learners that
  use little information,'' \emph{J. of Mach. Learn. Res.}, vol.~83, pp.
  25--55, Apr. 2018.

\bibitem{esposito19-12a}
\BIBentryALTinterwordspacing
A.~Esposito, M.~Gastpar, and I.~Issa, ``Generalization error bounds via
  {R{\`e}nyi} $f$-divergences and maximal leakage,'' \emph{arXiv}, Dec. 2019.
  [Online]. Available: \url{http://arxiv.org/abs/1912.01439}
\BIBentrySTDinterwordspacing

\bibitem{mcallester13-a}
\BIBentryALTinterwordspacing
D.~McAllester, ``A {PAC-B}ayesian tutorial with a dropout bound,'' July 2013.
  [Online]. Available: \url{http://arxiv.org/abs/1307.2118}
\BIBentrySTDinterwordspacing

\bibitem{YangRoy-2019}
J.~Yang, S.~Sun, and D.~M. Roy, ``Fast-rate {PAC-Bayes} generalization bounds
  via shifted {Rademacher} processes,'' in \emph{Proc. Conf. Neural Inf.
  Process. Syst. (NeurIPS)}, Vancouver, Canada, Dec. 2019.

\bibitem{Grunwald-20}
P.~Gr{\"{u}}nwald and N.~Mehta, ``Fast rates for general unbounded loss
  functions: from {ERM} to generalized {Bayes},'' \emph{J. of Mach. Learn.
  Res.}, vol.~83, pp. 1--80, Mar. 2020.

\bibitem{steinke20-a}
\BIBentryALTinterwordspacing
T.~Steinke and L.~Zakynthinou, ``Reasoning about generalization via conditional
  mutual information,'' in \emph{Proc. Conf. Learn. Theory (COLT)}, Graz,
  Austria, July 2020. [Online]. Available:
  \url{https://arxiv.org/abs/2001.09122}
\BIBentrySTDinterwordspacing

\bibitem{hellstrom-20b}
F.~{Hellström} and G.~{Durisi}, ``Generalization bounds via information
  density and conditional information density,'' \emph{IEEE J. Sel. Areas Inf.
  Theory}, vol.~1, no.~3, pp. 824--839, Dec. 2020.

\bibitem{dziugaite-20}
\BIBentryALTinterwordspacing
G.~Dziugaite, K.~Hsu, W.~Gharbieh, and D.~Roy, ``On the role of data in
  {PAC-Bayes} bounds,'' June 2020. [Online]. Available:
  \url{https://arxiv.org/abs/2006.10929}
\BIBentrySTDinterwordspacing

\bibitem{polyanskiy19-a}
\BIBentryALTinterwordspacing
Y.~Polyanskiy and Y.~Wu, \emph{Lecture Notes On Information Theory}, 2019.
  [Online]. Available:
  \url{http://www.stat.yale.edu/%7Eyw562/teaching/itlectures.pdf}
\BIBentrySTDinterwordspacing

\bibitem{Haghifam2020}
\BIBentryALTinterwordspacing
M.~Haghifam, J.~Negrea, A.~Khisti, D.~Roy, and G.~Dziugaite, ``Sharpened
  generalization bounds based on conditional mutual information and an
  application to noisy, iterative algorithms,'' \emph{arXiv}, Apr. 2020.
  [Online]. Available: \url{http://arxiv.org/abs/2004.12983}
\BIBentrySTDinterwordspacing

\bibitem{Dziugaite2017}
G.~Dziugaite and D.~Roy, ``Computing nonvacuous generalization bounds for deep
  (stochastic) neural networks with many more parameters than training data,''
  in \emph{Proc. Conf. Uncertainty in Artif. Intell. (UAI)}, Sydney, Australia,
  Aug. 2017.

\bibitem{wainwright19-a}
M.~J. Wainwright, \emph{High-Dimensional Statistics: a Non-Asymptotic
  Viewpoint}.\hskip 1em plus 0.5em minus 0.4em\relax Cambridge, U.K.: Cambridge
  Univ. Press, 2019.

\bibitem{zhou2018nonvacuous}
W.~Zhou, V.~Veitch, M.~Austern, R.~Adams, and P.~Orbanz, ``Non-vacuous
  generalization bounds at the {ImageNet} scale: a {PAC-B}ayesian compression
  approach,'' in \emph{Proc. Int. Conf. Learn. Representations (ICLR)}, New
  Orleans, LA, May 2019.

\end{thebibliography}

\appendices
\section{Proof of \eqref{eq:hellstrom_slow_pacb} and \eqref{eq:hellstrom_slow_sd}}\label{sec:app_A}
Let $(w,\supersamplesmall)$ be fixed, and consider the random variable $\Delta(\subsetchoice)= \testlosswz-\trainlosswz$. Due to the boundedness of $\ell(\cdot,\cdot)$ and the symmetry property $\Delta(\subsetchoice)=-\Delta(\bar \subsetchoice)$, $\Delta(\subsetchoice)$ is $1/\sqrt{n}$-sub-Gaussian with mean 0 under $P_{\subsetchoice}$. Applying \cite[Thm.~2.6.(IV)]{wainwright19-a} with $\lambda=1-1/n$, we conclude that
\begin{equation}
\Ex{P_{\subsetchoice} }{\exp\lefto(\frac{n-1}{2}(\testlosswz-\trainlosswz)^2 \right) } \leq \sqrt{n}.
\end{equation}

Taking the expectation with respect to $Q_{W\vert\supersample}P_{\supersample}$, changing measure to $\jointdistro$, and rearranging terms, we get
\begin{equation}\label{eq:slow_rate_exp_ineq}
    \Exop_{\jointdistro}\biggo[\exp\biggo(\frac{n-1}{2}(\testloss-\trainloss)^2-\log \sqrt{n} - \logQRN \bigg)\bigg] \leq 1.
\end{equation}
By similar steps as the ones used to prove Corollary~\ref{cor:bounds}, we obtain~\eqref{eq:hellstrom_slow_pacb}, after an additional use of Jensen's inequality, and~\eqref{eq:hellstrom_slow_sd}.

\section{Experiment Details}\label{app:experiment_details}

Here, we provide a detailed description of the network architectures and training procedures considered in this paper. 

\subsubsection{Network architectures}\label{app:exp_architectures}
The LeNet-5 architecture used in the numerical results is described in Table~\ref{tab:LeNet}. This is different from most standard implementations of LeNet-5, but coincides with the architecture used by~\cite{dziugaite-20} and~\cite{zhou2018nonvacuous}. It has~$431\,080$ parameters. Note that, for the binarized MNIST data set considered in Table~\ref{tab:randomized_labels}, the number of output units is instead $2$, resulting in a network with $427\,072$ parameters. The fully connected neural network denoted by~$600^2$ consists of an input layer with~$784$ units,~$2$ fully connected layers with~$600$ units and $\mathrm{ReLU}$ activations, followed by an output layer with~$10$ units and $\mathrm{softmax}$ activations. It has~$837\,610$ parameters. 
\begin{table}[ht]\caption{The LeNet-5 architecture used in Section~\ref{sec:experiments}.} \label{tab:LeNet}\centering
\begin{tabular}{l} 
\toprule 
Convolutional layer, 20 units, $5\times 5$ size, linear activation, $1\times 1$ stride, valid padding\\
Max pooling layer, $2\times 2$ size, $2\times 2$ stride \\
Convolutional layer, 50 units, $5\times 5$ size, linear activation, $1\times 1$ stride, valid padding\\
Max pooling layer, $2\times 2$ size, $2\times 2$ stride \\
Flattening layer \\
Fully connected layer, 500 units,~$\mathrm{ReLU}$ activation \\
Fully connected layer, 10 units, $\mathrm{softmax}$ activation
\\ \bottomrule
\end{tabular}\end{table}%
\subsubsection{Training procedures}\label{app:exp_training}
We now provide additional details on the training procedures described in Section~\ref{sec:experiments}. The initial weights of all the networks used for each instance of~$\trainingdata(\subsetchoice)$ were set to the same randomly selected values drawn from a zero-mean normal distribution with standard deviation~$0.01$.
All networks were trained using the cross-entropy loss, optimized using either SGD with momentum and a fixed learning rate or SGD without momentum and a decaying learning rate. First, we describe the details of SGD with momentum. For MNIST, we used a learning rate of~$0.001$, and for Fashion-MNIST, we used~$0.003$. In all experiments, the momentum parameter is set to~$0.9$. We used a batch size of~$512$.

For SGD without momentum, we used a decaying learning rate schedule, where the learning rate~$\alpha$ for a given epoch~$E$ is given by
\begin{equation}\label{eq:learning_rate_decay}
  \alpha(E)  = \frac{\alpha_0}{ 1+\gamma \cdot \lfloor E/E_0 \rfloor }.
\end{equation}
Here,~$\alpha_0$ is the initial learning rate,~$\gamma$ is the decay rate, and~$E_0$ is the number of epochs between each decay. 
In all experiments, we used~$\alpha_0=0.01$,~$\gamma=2$, and~$E_0=20$. Again, we used a batch size of $512$.

To choose~$\sigma_1$, we pick the largest value with one significant digit (i.e., of the form~$a\cdot 10^{-b}$ with~$a\in [1:9]$ and~$b\in \integers$) such that the absolute value of the difference between the training loss on~$\trainingdata(\subsetchoice)$ of the deterministic network with weights~$\boldsymbol{\mu}_1$ and empirical average of the training loss of $5$ NNs with weights drawn independently from~$\normal(W\mid \boldsymbol{\mu}_1,\sigma^2_1 \matI_d)$ was no larger than some specified threshold. 

When producing the results reported in Figures~\ref{fig:plots_part_1}a--c, \ref{fig:plots_part_1}e--f for MNIST, we use a threshold of $0.05$. For Figures \ref{fig:plots_part_2}a--c, \ref{fig:plots_part_2}e--f, where we consider Fashion-MNIST, we use a threshold of~$0.10$. 
For Figure~\ref{fig:plots_part_1}d, we use a threshold of $0.15$, while we use a threshold of $0.45$ for Figure~\ref{fig:plots_part_2}d. For the randomized label experiment in Table~\ref{tab:randomized_labels}, we use a threshold of $0.10$.

Next, $\boldsymbol{\mu}_2$ is determined as follows. We form~$10$ subsets of~$\supersample$, each of size~$n$. 
The first subset contains the first~$n$ samples in~$\supersample$, the last contains the last~$n$ samples in~$\supersample$, and the remaining subsets contain the linearly spaced sequences in between. 
We then train one NN on each subset and denote the average of the final weights of these networks by~$\boldsymbol{\mu}_2$. 
To find~$\sigma_2$, we use as starting point the same procedure as for determining~$\sigma_1$, but with~$\boldsymbol{\mu}_2$ in place of~$\boldsymbol{\mu}_1$ and the training loss evaluated on all of~$\supersample$. 
Let us call the value found by this procedure~$\sigma_2'=a'\cdot 10^{-b'}$. Then, among the values of the form~$a\cdot 10^{-b}$ with~$a\in [1:9]$ and~$b\in \{b'-1,b',b'+1\}$, we choose~$\sigma_2$ to be the one that minimizes the bound on the test loss. 
In all our experiments, this procedure resulted in~$\sigma_2=\sigma_1$. 
To guarantee that the final bound holds with a given confidence level, all~$27$ bounds resulting from all possible choices of $a$ and $b$ need to hold with the same confidence level. 
Since we consider both slow-rate and fast-rate bounds, a total of~$54$ bounds need to hold simultaneously. 
We ensure that this is the case via the union bound. 
Thus, if each individual bound holds with probability at least~$1-\delta$, the optimized bounds hold with probability at least~$1-54 \delta$. 
We compute the bounds with~$\delta = 0.05/54$, so the optimized bounds hold with~$95\%$ confidence.
\end{document}